\pdfoutput=1

\documentclass[11pt]{article}

\usepackage{acl}

\usepackage{times}
\usepackage{latexsym}
\usepackage{microtype}
\usepackage{makecell}
\usepackage{hyperref}
\usepackage{graphicx}
\usepackage{multirow}
\usepackage{amsmath}
\usepackage{xcolor}
\usepackage{booktabs}
\usepackage{caption}
\usepackage{subcaption}
\usepackage{amsfonts}
\usepackage{tabularx}
\usepackage{rotating}
\usepackage[T1]{fontenc}
\usepackage[utf8]{inputenc}
\usepackage{microtype}
\usepackage{inconsolata}
\usepackage{graphicx}
\usepackage{amssymb}
\usepackage{algorithmicx}
\usepackage{algorithm}
\usepackage{algpseudocode}
\usepackage{amsmath}
\algnewcommand\algorithmicendwhile{\textbf{end\ while}}
\algrenewtext{While}[1]{\algorithmicwhile\ #1\ \algorithmicdo}
\usepackage{caption}

\usepackage{dashrule} 

\newcommand{\dashedline}{\textcolor{blue}{\hdashrule{0.5cm}{0.4pt}{0.5mm}}}
\newcommand{\orangeline}{\textcolor{orange}{\rule{0.5cm}{0.4pt}}}
\newcommand{\greenline}{\textcolor{green}{\rule{0.5cm}{0.4pt}}}

\definecolor{customred}{HTML}{ff999a}   
\definecolor{customgray}{HTML}{7f7f7f}  
\definecolor{customblue}{HTML}{b5c7e7}  
\newcommand{\redrect}{\colorbox{customred}{\phantom{\rule{0.5cm}{0.15cm}}}} 
\newcommand{\grayrect}{\colorbox{customgray}{\phantom{\rule{0.5cm}{0.15cm}}}} 
\newcommand{\bluerect}{\colorbox{customblue}{\phantom{\rule{0.5cm}{0.15cm}}}} 

\title{Mitigating Negative Interference in Multilingual Sequential Knowledge Editing through Null-Space Constraints}

\author{Wei Sun, Tingyu Qu,  Mingxiao Li\thanks{\ \ Corresponding author.}, Jesse Davis \and Marie-Francine Moens \\
        Department of Computer Science, KU Leuven \\
        Celestijnenlaan 200A 3001 Heverlee, Belgium \\
        \texttt{\{sun.wei, tingyu.qu, mingxiao.li, jesse.davis, sien.moens\}@kuleuven.be}}

\begin{document}
\maketitle
\begin{abstract}

Efficiently updating multilingual knowledge in large language models (LLMs), while preserving consistent factual representations across languages, remains a long-standing and unresolved challenge. 
While deploying separate editing systems for each language might seem viable, this approach incurs substantial costs due to the need to manage multiple models. A more efficient solution involves integrating knowledge updates across all languages into a unified model. 
However, performing sequential edits across languages often leads to destructive parameter interference, significantly degrading multilingual generalization and the accuracy of injected knowledge.
To address this challenge, we propose LangEdit, a novel null-space constrained framework designed to precisely isolate language-specific knowledge updates. 
The core innovation of {LangEdit} lies in its ability to project parameter updates for each language onto the orthogonal complement of previous 
updated subspaces. This approach mathematically guarantees update independence while preserving multilingual generalization capabilities.
We conduct a comprehensive evaluation across three model architectures, six languages, and four downstream tasks, demonstrating that {LangEdit} effectively mitigates parameter interference and outperforms existing state-of-the-art editing methods. 
Our results highlight its potential for enabling efficient and accurate multilingual knowledge updates in LLMs.
The code is available at \url{https://github.com/VRCMF/LangEdit.git}. 
\end{abstract}

\section{Introduction}

\begin{figure}[!t]
\centering
\centering
    \begin{minipage}{0.5\textwidth} %
        \centering
        \subfloat[\scriptsize{Monolingual Knowledge Editing (multiple edited models).}]{
            \includegraphics[width=0.75\textwidth]{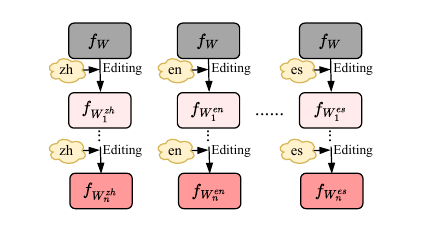}
            \label{fig:intro_mono}
        }
        \hfil
        \subfloat[\scriptsize{Multilingual Knowledge Editing (only one edited model).}]{
            \includegraphics[width=1\textwidth]{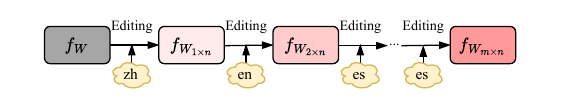}
            \label{fig:intro_demo}
        }
    \end{minipage}
\caption{
Illustration of Monolingual (a) and Multilingual (b) Sequential Knowledge Editing.
\( f_W \) denote the pre-edited model. 
After performing sequential edits, \( f_W^* \) represents the set of monolingual models, where each model has undergone \( n \) edits, and its superscript indicates the specific language of editing.
For multilingual scenarios, \( f_{W_{m \times n}} \) represents a single multilingual model trained on \( m \) languages, with each language containing \( n \) samples.
}
\label{fig:intro_illu}
\end{figure}

Modern large language models (LLMs) exhibit strong capabilities in encoding and retrieving factual knowledge. 
Knowledge editing has emerged as an efficient approach to updating knowledge within LLMs, reducing hallucinations without the need for resource-intensive retraining~\cite{gu-etal-2024-model}. 
However, efficiently updating knowledge in multilingual scenarios remains a significant challenge, as models must maintain factual consistency across multiple languages. 
Although existing monolingual knowledge editing methods~\cite{mengmass,gu-etal-2024-model,fang2024alphaedit,ma2024perturbation} have shown promising results, they lack effective solutions for managing multiple monolingual models concurrently, as illustrated in Figure~\ref{fig:intro_mono}.

To address this challenge, we introduce the task of multilingual sequential knowledge editing, which involves updating knowledge across multiple languages in a sequential manner, as depicted in Figure~\ref{fig:intro_demo}. 
This task holds particular significance for applications including knowledge-informed multilingual information retrieval~\cite{zhang2022mind,wang2024cl2cm} and factual updates of multilingual LLMs~\cite{singhal-etal-2024-multilingual}.
A critical challenge arises from the fact that editing knowledge in one language can negatively impact the model's performance in other languages.
We term this phenomenon \textit{negative interference}, describing how knowledge editing in a multilingual setting degrades performance on previously updated languages and undermines the multilingual generalization capabilities of LLMs, as demonstrated in our experiments.

We propose {LangEdit}, a multilingual knowledge editing framework designed to mitigate negative interference. 
LangEdit constrains parameter updates for each language to the null space \cite{greub2012linear}  
of previous language updated subspaces. For example, when editing Chinese facts after English updates, LangEdit projects Chinese updates onto the null-space of the parameter updates for English, mathematically ensuring minimal negative interference. This approach creates protective "language safeguards" that prevent parameter conflicts. 
Compared to existing editing models, LangEdit yields substantial improvements, achieving up to 5.65 percentage points increase in multilingual generalization tasks and 2.20 percentage points improvement in editing accuracy, thereby demonstrating its dual capacity for precise knowledge editing and effective multilingual knowledge retention.

We conduct extensive experiments spanning three model architectures, six languages, four multilingual generalization evaluation tasks, and two multilingual knowledge editing datasets. Across these settings, LangEdit consistently outperforms strong sequential editing baselines, establishing state-of-the-art performance. 
Our contributions are:
\begin{itemize}
    \item We introduce the task of multilingual sequential knowledge editing, which involves sequentially updating a multilingual LLM with knowledge in multiple languages. 
    \item We develop LangEdit, a null-space projection framework that 
    provably perform language-specific knowledge updates.
    \item We show LangEdit's consistent gains across diverse languages, model scales and downstream tasks commonly used for evaluating the multilingual generalization ability.
\end{itemize}

\section{Preliminaries}

\subsection{Multilingual Knowledge Editing in LLMs}

An autoregressive LLM iteratively generates the next token of a sentence based on the previously generated tokens. 
Let \( h^l \) denote the hidden state of the next token at layer \( l \). 
The hidden state is computed as follows~\cite{meng2022locating,fang2024alphaedit}: 
: 

\begin{equation}
\begin{aligned}
\mathbf{h}^l &= \mathbf{h}^{l-1} + \mathbf{a}^l + \mathbf{m}^l, \\
\mathbf{m}^l &= \mathbf{W}_{\text{out}}^{l} \sigma(\mathbf{W}_{\text{in}}^{l} \gamma(\mathbf{h}^{l-1} + \mathbf{a}^l)),
\end{aligned}
\label{eq:llm_eq}
\end{equation}
where \( \mathbf{a}^l \) represents the output of the attention block in layer $l$ and \( \mathbf{m}^l \) corresponds to the output of the multilayer perceptron (MLP) in layer $l$.
\( \mathbf{W}_{\text{out}}^{l} \) and \( \mathbf{W}_{\text{in}}^{l} \) are weight matrices of the $l_{th}$ MLP.
\( \sigma \) is the activation function and \( \gamma \) represents layer normalization. 
Following prior works~\cite{meng2022locating,mengmass,fang2024alphaedit}, we express the attention block and MLP in parallel. 

The MLP layers can be interpreted as linear associative memory~\cite{geva-etal-2021-transformer}.
Specifically, the knowledge stored in the model can be formalized as triplets~($s$, $r$, $o$), where $s$ represents the subject, $r$ the relation, and $o$ the object.
For example, the triplet ($s$ = Space Needle, $r$ = is in,
$o$ = the center of Seattle) encodes a factual relationship.
In this framework, the output representation of $\sigma(\mathbf{W}_{\text{in}}^{l} \gamma(\mathbf{h}^{l-1} + \mathbf{a}^l))$ corresponds to the key~(subject and relation) of the knowledge, while $\mathbf{m}^l$ represents the value~(object) of the knowledge.
In the multilingual sequential knowledge editing, our objective is to optimize \( \mathbf{W}_{out}^l \) within the neural network \( f \).
For simplicity of notation, in what follows we leave out these sub- and superscripts of $\mathbf{W}_{\text{out}}^{l}$ and write $\mathbf{W}$.

Let the stacked keys of new input knowledge in language $j$ at time step $t$ be $\mathbf{K}_{t} = [\mathbf{k}_{t,1} \mid \mathbf{k}_{t,2} \mid \ldots \mid \mathbf{k}_{t,n_t}] \in \mathbb{R}^{d_0 \times n_t}$, where $n_t$ denotes the number of keys and $d_0$ is the dimension of the intermediate layer.
Note that the language index $j$ is omitted in the following equations for simplicity. 
Similarly, let the corresponding values be $ \mathbf{V}_{t} = [\mathbf{v}_{t,1} \mid \mathbf{v}_{t,2} \mid \ldots \mid \mathbf{v}_{t,n_t}] \in \mathbb{R}^{d_1 \times n_t}$, where $d_1$
is the dimension of the output layer.
Each input data \( \mathbf{L}_t \) is represented as a set of key-value pairs \(\{\mathbf{K}_t, \mathbf{V}_t\}\) with a size of \( n_t \). 
The network is trained sequentially on a stream of multilingual data \( \{\mathbf{L}_1, \mathbf{L}_2, \cdots, \mathbf{L}_T\} \) , where each data \( \mathbf{L}_t \) corresponds to the time step \( t \) that introduces knowledge of a language $j$. 
The initial parameters for training on \( \mathbf{L}_t \) are initialized as \( {\mathbf{W}}_{t-1} \), which are the optimal parameters obtained after training on the previous data \( \mathbf{L}_{t-1} \). 

\subsection{Null-space Projection}

Assume that the model \(f\) is trained on data \(\mathbf{L}_t\) (for \(t > 1\)), with the corresponding weight update denoted as \(\mathbf{\Delta} \mathbf{W}_{t}\).
Let \(\bar{\mathbf{K}}_{t-1} = [\mathbf{K}_{1}; \cdots; \mathbf{K}_{t-1}]\) and \(\bar{\mathbf{V}}_{t-1} = [\mathbf{V}_{1}; \cdots; \mathbf{V}_{t-1}]\) represent the concatenated keys and values from all previous update steps, respectively.
To mitigate the caused negative interference, 
the updates are constrained to 
lie within the null space of the previously injected knowledge representations.
Specifically, the weight updates \(\mathbf{\Delta} \mathbf{W}_{t}\) are projected into the null space of \(\bar{\mathbf{K}}_{t-1}\) and compute:
\begin{equation}
(\mathbf{\Delta} \mathbf{W}_{t} + \mathbf{W}_{t-1})\bar{\mathbf{K}}_{t-1} = \bar{\mathbf{V}}_{t-1}, 
\label{eq:preserve_old_knowldege}
\end{equation}

As the dimensions of \(\bar{\mathbf{K}}_{t-1}\) and \(\bar{\mathbf{V}}_{t-1}\) grow with the injection of additional knowledge, computing the null space becomes computationally expensive.
To address this, we replace the concatenated key and value matrices with their corresponding uncentered covariance matrices, inspired by techniques used in continual learning for computer vision~\cite{wang2021training}:
\(
\bar{\boldsymbol{\mathcal{K}}}_{t-1} \triangleq \frac{1}{\bar{n}_{t-1}}(\bar{\mathbf{K}}_{t-1})^\top \bar{\mathbf{K}}_{t-1},
\)
and
\(
\bar{\boldsymbol{\mathcal{V}}}_{t-1} \triangleq \frac{1}{\bar{n}_{t-1}}(\bar{\mathbf{V}}_{t-1})^\top \bar{\mathbf{V}}_{t-1},
\)
where \(\bar{n}_{t-1}\) is the number of rows in \(\bar{\mathbf{K}}_{t-1}\). 
Consequently, Equation~\ref{eq:preserve_old_knowldege} is reformulated as: 
\begin{equation}
(\mathbf{\Delta} \mathbf{W}_{t} + \mathbf{W}_{t-1})\bar{\boldsymbol{\mathcal{K}}}_{t-1} = \bar{\boldsymbol{\mathcal{V}}}_{t-1}. 
\label{eq:preserve_old_knowldege_update}
\end{equation}
It is straightforward to verify that the null space of \(\bar{\mathbf{K}}_{t-1}\) is equivalent to the null space of the uncentered feature covariance matrix~\(\bar{\boldsymbol{\mathcal{K}}}_{t-1}\).

\section{Method}

This section presents \textbf{LangEdit}, a novel model designed for multilingual sequential knowledge editing. 
Our approach utilizes feature covariance matrices, which are incrementally updated as new language-specific data arrives. 
Specifically, given a network trained on prior multilingual data, LangEdit updates the model parameters corresponding to language $j$ at the current time step $t$.
The key innovation lies in projecting the candidate parameter updates into the approximate null space of the feature covariance matrix of the previously learned knowledge representation, ensuring that
language-specific knowledge updates remain
decoupled and do not interfere with each other. 

After training on data \( \mathbf{L}_{t-1} \), we compute the uncentered covariance matrix as follows: 
\(
{\boldsymbol{\mathcal{K}}}_{t-1} = \frac{1}{{n}_{t-1}}({\mathbf{K}}_{t-1})^\top {\mathbf{K}}_{t-1},
\) 
where \( n_{t-1} \) denotes the number of data points in \( \mathbf{L}_{t-1} \).
If the \( \mathbf{L}_{t} \) is in language $j$, the uncentered feature covariance matrix is then updated recursively:
\begin{equation}
\bar{\boldsymbol{\mathcal{K}}}_{t-1} = \frac{\bar{n}_{t-2}}{\bar{n}_{t-1}} \bar{\boldsymbol{\mathcal{K}}}_{t-2} + \frac{n_{t-1}}{\bar{n}_{t-1}} \boldsymbol{\mathcal{K}}_{t-1},
\end{equation}
where \( \bar{n}_{t-1} = \bar{n}_{t-2} + n_{t-1} \) represents the cumulative number of data points observed up to step \( t-1 \). 
To preserve previously injected knowledge, we compute the approximate null space of \( \bar{\boldsymbol{\mathcal{K}}}_{t-1} \) (Equation~\ref{eq:preserve_old_knowldege_update}).

When training the network with \( {\mathbf{W}}_{t-1} \) as initialization on data \( \mathbf{L}_t \), the parameter update $\Delta \mathbf{W}_{t}$ is obtained by optimizing the following objective:
\begin{equation}
\begin{split}
\Delta \mathbf{W}_{t} = \arg \min_{\mathbf{\tilde{\Delta \mathbf{W}_{t}}}} \left( \left\| (\mathbf{\tilde{\Delta \mathbf{W}_{t}}} + \mathbf{W}_{t-1}) \mathbf{K}_t - \mathbf{V}_t \right\|^2 \right. \\
\left. + \left\| (\mathbf{\tilde{\Delta \mathbf{W}_{t}}} + \mathbf{W}_{t-1}) \bar{\mathbf{K}}_{t-1} - \bar{\mathbf{V}}_{t-1} \right\|^2 \right),
\end{split}
\label{eq:obj_ori}
\end{equation}
where \(\mathbf{\tilde{\Delta \mathbf{W}_{t}}}\)  is the optimization variable. 

To ensure that parameter updates reside in the null space of the uncentered covariance matrix of previous data, we adopt the methodology described in~\cite{wang2021training,fang2024alphaedit}.
Specifically, we construct an approximate null space by performing Singular Value Decomposition~(SVD) of $\bar{\boldsymbol{\mathcal{K}}}_{t-1}$: 
\begin{equation}
    \mathbf{U}_{t-1}, \boldsymbol{\Sigma}_{t-1}, \mathbf{V}_{t-1} = \operatorname{SVD}(\bar{\boldsymbol{\mathcal{K}}}_{t-1}),
\end{equation}
We retain only eigenvectors in $\mathbf{U}_{t-1}$ corresponding to zero eigenvalues, yielding the matrix $\mathbf{U}_{t-1}'$.
The null space projection matrix is then defined as:
\begin{equation}
\mathbf{P}_{t-1} = \mathbf{U}_{t-1}' (\mathbf{U}_{t-1}')^{\top}
\end{equation}

The parameter update 
\(\mathbf{\tilde{\Delta \mathbf{W}_{t}}} \)
is projected onto the null space of \(\bar{\boldsymbol{\mathcal{K}}}_{t-1}\) using $\mathbf{P}_{t-1}$, the right hand side of Equation~\ref{eq:obj_ori} becomes:
\begin{equation}
\arg \min_{\mathbf{\tilde{\Delta \mathbf{W}_{t}}}} \left\| (\mathbf{P}_{t-1} \mathbf{\tilde{\Delta \mathbf{W}_{t}}} + \mathbf{W}_{t-1}) \mathbf{K}_t - \mathbf{V}_t \right\|^2,
\label{eq:projected_obj}
\end{equation}

because \((\mathbf{P}_{t-1} \mathbf{\tilde{\Delta \mathbf{W}_{t}}} + \mathbf{W}_{t-1})\bar{\mathbf{K}}_{t-1} = \bar{\mathbf{V}}_{t-1}, \) where \(\mathbf{P}_{t-1}\) is the null space projection matrix. 
To stabilize convergence, a regularization term \(\mathbf{P}_{t-1} \mathbf{\tilde{\Delta \mathbf{W}_{t}}}\) is added, leading to the final optimization objective:
\begin{equation}
\begin{split}
\Delta \mathbf{W}_{t} = \arg \min_{\mathbf{\tilde{\Delta \mathbf{W}_{t}}}} \left( 
\left\| \mathbf{P}_{t-1} \mathbf{\tilde{\Delta \mathbf{W}_{t}}} \right\|^2 \right. \\
+ \left. \left\| (\mathbf{P}_{t-1} \mathbf{\tilde{\Delta \mathbf{W}_{t}}} + \mathbf{W}_{t-1}) \mathbf{K}_t - \mathbf{V}_t \right\|^2 \right), 
\end{split}
\label{eq:obj}
\end{equation}
The closed solution for the final objective is: 
\begin{equation}
\Delta \mathbf{W}_t = \mathbf{R}_t  \mathbf{K}_t^\top\mathbf{P}_{t-1}( \mathbf{K}_t \mathbf{K}_t^\top \mathbf{P}_{t-1} + \mathbf{I})^{-1}.
\end{equation}
where $\mathbf{R}_t = (\mathbf{V}_t - \mathbf{W}_{t-1} \mathbf{K}_t)$.
A critical step in this process is computing the stacked keys $\mathbf{K}_0$, which represent the old knowledge stored in the LLM.
Following the approach in~\cite{meng2022locating}, $\mathbf{K}_0 \in \mathbb{R}^{d_0 \times 100,000}$ is computed using randomly sampled triplets from Wikipedia. 
This procedure is consistently applied across all models and baselines.
The LangEdit method is summarized in Algorithm~\ref{alg:langedit}.

\begin{algorithm}[t]
\caption{LangEdit: multilingual sequential knowledge editing}
\label{alg:langedit}
\begin{algorithmic}[1]
\Require Initialized weight $\mathbf{W}_0$, Data sequence $\{\mathbf{L}_1, \dots, \mathbf{L}_T\}$, Initial covariance $\bar{\boldsymbol{\mathcal{K}}}_{0} = \boldsymbol{\mathcal{K}}_0 =  \frac{1}{{n}_{0}}\mathbf{K}_0^\top \mathbf{K}_0$, Data sizes $\{n_t\}_{t=1}^{T}$
\For{$t = 1$ to $T$}
    \State Extract input keys: $\mathbf{K}_t$
    \State Extract corresponding values: $\mathbf{V}_t$
    \State Obtain previous keys: ${\mathbf{K}}_{t-1}$
    \State Compute covariance: $
    {\boldsymbol{\mathcal{K}}}_{t-1} \gets \frac{1}{{n}_{t-1}}({\mathbf{K}}_{t-1})^\top {\mathbf{K}}_{t-1},
    $
    \State if $t > 1$, then update running covariance: $\bar{\boldsymbol{\mathcal{K}}}_{t-1} \gets \frac{\bar{n}_{t-2}}{\bar{n}_{t-1}} \bar{\boldsymbol{\mathcal{K}}}_{t-2} + \frac{n_{t-1}}{\bar{n}_{t-1}} \boldsymbol{\mathcal{K}}_{t-1}$, $\bar{n}_{t-1} = \bar{n}_{t-2} + n_{t-1}$
    \State Perform SVD: $(\mathbf{U}_{t-1}, \boldsymbol{\Sigma}_{t-1}, \mathbf{V}_{t-1}) \gets \text{SVD}(\bar{\boldsymbol{\mathcal{K}}}_{t-1})$
    \State Compute projection matrix: $\mathbf{P}_{t-1} \gets \mathbf{U}'_{t-1} \mathbf{U'}_{t-1}^\top$ (Keep eigenvectors in $\mathbf{U}_{t-1}$ whose eigenvalues are zero, to obtain $\mathbf{U}'_{t-1}$).
    \State Solve optimization:
    \[
            \begin{array}{l}
    \Delta\mathbf{W}_t = \arg\min_{\tilde{\mathbf{\Delta}\mathbf{W_t}}} \Big( 
    \|\mathbf{P}_{t-1} \tilde{\mathbf{\Delta}\mathbf{W_t}} \|^2  + \\
    \quad \| (\mathbf{P}_{t-1} \tilde{\mathbf{\Delta}\mathbf{W_t}} + \mathbf{W}_{t-1}) \mathbf{K}_t - \mathbf{V}_t \|^2 \Big)
    \end{array}
        \]
    \State Update model: $\mathbf{W}_t \gets \mathbf{W}_{t-1} + \Delta\mathbf{W}_t$
\EndFor
\Ensure Updated model $\mathbf{W}_T$
\end{algorithmic}
\end{algorithm}

\section{Experiments}

\subsection{Experimental Setup}

\subsubsection{Datasets}

Since multilingual sequential knowledge editing is a novel task, 
we constructed a benchmark using two multilingual datasets\cite{wang-etal-2024-cross,wang-etal-2024-retrieval}.
We evaluated all models and baselines on these datasets.
\textbf{bzsre:} The Bi-ZsRE(Bilingual Zero-Shot Relation Extraction) dataset~\cite{wang-etal-2024-cross} is designed to assess the impact of knowledge editing in multilingual LLMs.
It comprises question-answer pairs in two languages: English and Chinese.
From this dataset, we randomly selected 800 samples per language to create a bilingual sequential knowledge editing dataset (bzsre).
The total number of edits is 1600.
\textbf{mzsre:} The Multilingual Zero-Shot Relation Extraction (M-ZsRE) dataset~\cite{wang-etal-2024-retrieval} contains question-answer pairs in twelve languages. 
Due to constraints of the downstream task, we focused on six languages: English, German, Dutch, Spanish, French, and Chinese.
For each language, we extracted 400 samples to construct a multilingual sequential knowledge editing dataset (mzsre).
The total number of edits is 2400.
We put the description for MLaKE dataset~\cite{wei-etal-2025-mlake} and experimental results on this dataset to the Appendix~\ref{sec:mlake}. 

To evaluate the multilingual generalization capabilities of the edited LLMs, we employed four tasks from the Cross-lingual TRansfer Evaluation of Multilingual Encoders~(XTREME) benchmark~\cite{hu2020xtreme}:  
\textbf{XNLI} A cross-lingual natural language inference (NLI) benchmark extending MultiNLI~\cite{gururangan-etal-2018-annotation} to 15 languages. It evaluates multilingual generalization in NLI tasks through textual entailment classification.
\textbf{PAWS-X}~\cite{yang-etal-2019-paws} A cross-lingual paraphrase identification dataset featuring adversarial examples generated via word-order perturbations. It tests model robustness against structural ambiguities in multilingual contexts.
\textbf{MLQA}~\cite{lewis-etal-2020-mlqa} A multilingual question answering benchmark derived from Wikipedia articles. It measures extractive question-answering (QA) performance across 7 languages.
\textbf{Wikiann}~\cite{pan-etal-2017-cross} A multilingual named entity recognition resource with consistent PER/ORG/LOC annotations, providing standardized evaluation for multilingual named-entity recognition (NER) model adaptation.

\subsubsection{Metrics}
Following prior works \citep{meng2022locating,mengmass,fang2024alphaedit}, we define each editing metric given a LLM \( f \), a knowledge fact prompt \((s_i, r_i)\), the target output of the edited model \( o_i \), and the output of the original model \( o_i^c \) as follows:

\noindent \textbf{Efficacy}: Efficacy quantifies the model's ability to produce the target object $o_i$ when prompted with $(s_i, r_i)$. It is computed as the average top-1 accuracy over all edited samples:
\begin{equation}
\mathbb{E}_i \left\{ o_i = \arg \max_o \mathbb{P}_{f} (o \mid (s_i, r_i)) \right\}. 
\label{eq:efficacy_metric}
\end{equation}
\textbf{Generality}: Generality evaluates the performance of the model on equivalent prompts of \((s_i, r_i)\), such as rephrased statements \( N((s_i, r_i)) \). This is evaluated by the average top-1 accuracy on these \( N((s_i, r_i)) \):
\begin{equation}
\mathbb{E}_i \left\{ o_i = \arg \max_o \mathbb{P}_{f} (o \mid N((s_i, r_i))) \right\}. 
\label{eq:generalization_metric}
\end{equation}
\textbf{Specificity}: Specificity ensures that the editing does not affect samples \( O(s_i, r_i) \) which are unrelated to the edit cases. This is evaluated by the top-1 accuracy of predictions that should remain unchanged:
\begin{equation}
\mathbb{E}_i \left\{ o_i^c = \arg \max_o \mathbb{P}_{f} (o \mid O((s_i, r_i))) \right\}. 
\label{eq:specificity_metric}
\end{equation}

To evaluate multilingual generalization in edited LLMs, we report \textbf{F1} Scores across four NLP tasks in the XTREME benchmark. \textbf{F1~(avg.)} denotes the average F1 Score across these tasks. 
\begin{table*}[t!]
    \centering
    {\small
    \renewcommand{\arraystretch}{1.2}
    \setlength{\tabcolsep}{2pt}
    \begin{tabular}{ll |lll|l|lll|l} 
        \toprule
        \textbf{Model} & \multicolumn{1}{c}{\textbf{Methods}} & 
        \multicolumn{3}{c}{\textbf{mzsre}} & XTREME$\ddag$ & \multicolumn{3}{c}{\textbf{bzsre}} & XTREME$\dag$ \\
        \cmidrule(lr){3-5} \cmidrule(lr){7-9}
         &\multicolumn{1}{c}{} & \multicolumn{1}{c}
         {\textbf{Efficacy$\uparrow$}} & \textbf{Generality$\uparrow$} & \textbf{Specificity$\uparrow$} & {\textbf{F1~(avg.)$\uparrow$}} & {\textbf{Efficacy$\uparrow$}} & \textbf{Generality$\uparrow$} & \textbf{Specificity$\uparrow$} & \textbf{F1~(avg.)$\uparrow$} \\
        \midrule
        \multirow{8}{*}{\rotatebox{90}{Llama3-8B}} & Pre-edited & 31.15\textsubscript{$\pm$0.13} & 31.01\textsubscript{$\pm$0.22} & 31.93\textsubscript{$\pm$0.17} & 69.3\textsubscript{$\pm$0.21} & 31.55\textsubscript{$\pm$0.11} & 31.17\textsubscript{$\pm$0.15} & 30.5\textsubscript{$\pm$0.19} & 69.38\textsubscript{$\pm$0.25} \\
        \hline
        & FT & 30.76\textsubscript{$\pm$0.12} & 29.48\textsubscript{$\pm$0.19} & 9.53\textsubscript{$\pm$0.11} & 10.57\textsubscript{$\pm$0.03} & 31.41\textsubscript{$\pm$0.15} & 29.97\textsubscript{$\pm$0.17} & 15.29\textsubscript{$\pm$0.21} & 44.38\textsubscript{$\pm$0.33} \\
        & ROME & 0.39\textsubscript{$\pm$0.16} & 0.35\textsubscript{$\pm$0.18} & 0.32\textsubscript{$\pm$0.22} & 4.51\textsubscript{$\pm$0.01} & 2.54\textsubscript{$\pm$0.42} & 2.46\textsubscript{$\pm$0.44} & 0.39\textsubscript{$\pm$0.49} & 4.66\textsubscript{$\pm$0.19} \\
        & MEMIT & 
        1.45\textsubscript{$\pm$1.03} & 1.46\textsubscript{$\pm$0.99} & 0.67\textsubscript{$\pm$0.03} & 4.54\textsubscript{$\pm$0.00} & 4.58\textsubscript{$\pm$0.27} & 4.03\textsubscript{$\pm$0.09} & 2.84\textsubscript{$\pm$0.42} & 15.64\textsubscript{$\pm$2.41} \\
        & PRUNE & 0.43\textsubscript{$\pm$0.23} & 0.36\textsubscript{$\pm$0.02} & 0.02\textsubscript{$\pm$0.01} & 4.48\textsubscript{$\pm$0.00} & 4.92\textsubscript{$\pm$1.53} & 
        4.22\textsubscript{$\pm$1.23} & 1.90\textsubscript{$\pm$0.80} & 9.32\textsubscript{$\pm$2.14} \\
        & RECT & 2.91\textsubscript{$\pm$0.38} & 2.79\textsubscript{$\pm$0.36} & 0.64\textsubscript{$\pm$0.18} & 7.71\textsubscript{$\pm$2.49} & 41.01\textsubscript{$\pm$2.47} & 38.58\textsubscript{$\pm$1.95} & 20.80\textsubscript{$\pm$0.76} & 55.97\textsubscript{$\pm$3.44} \\
        & AlphaEdit & \underline{80.34}\textsubscript{$\pm$0.58} & \underline{75.84}\textsubscript{$\pm$0.73} & \underline{30.91}\textsubscript{$\pm$0.49} & \underline{60.59}\textsubscript{$\pm$0.38} & \underline{71.88}\textsubscript{$\pm$0.81} & \underline{66.55}\textsubscript{$\pm$0.33} & \underline{30.47}\textsubscript{$\pm$0.20} & \underline{71.25}\textsubscript{$\pm$0.70} \\
        & LangEdit & \textbf{82.54}\textsubscript{$\pm$0.14}*  & \textbf{77.53}\textsubscript{$\pm$0.43}* & \textbf{31.90}\textsubscript{$\pm$0.14}* & \textbf{66.24}\textsubscript{$\pm$0.28}* & \textbf{73.18}\textsubscript{$\pm$0.35}* & \textbf{66.95}\textsubscript{$\pm$0.17}* & \textbf{31.11}\textsubscript{$\pm$0.18}* & \textbf{73.14}\textsubscript{$\pm$0.83}* \\
        \bottomrule
        \multirow{8}{*}{\rotatebox{90}{Qwen2.5-7B}} & Pre-edited & 33.52\textsubscript{$\pm$0.09} & 33.11\textsubscript{$\pm$0.13} & 38.76\textsubscript{$\pm$0.08} & 71.9\textsubscript{$\pm$0.15} & 33.77\textsubscript{$\pm$0.07} & 33.24\textsubscript{$\pm$0.20} & 38.81\textsubscript{$\pm$0.15} & 73.91\textsubscript{$\pm$0.14} \\
        \hline
        & FT & 32.46\textsubscript{$\pm$0.13} & 30.28\textsubscript{$\pm$0.47} & 28.76\textsubscript{$\pm$0.20} & 45.20\textsubscript{$\pm$0.40} & 35.6\textsubscript{$\pm$0.15} & 33.06\textsubscript{$\pm$0.23} & 33.48\textsubscript{$\pm$0.17} & 59.51\textsubscript{$\pm$0.43} \\
        & ROME & 12.44\textsubscript{$\pm$0.47} & 11.35\textsubscript{$\pm$0.73} & 2.25\textsubscript{$\pm$0.71} & 4.70\textsubscript{$\pm$0.15} & 16.36\textsubscript{$\pm$0.77} & 15.27\textsubscript{$\pm$0.53} & 1.60\textsubscript{$\pm$0.29} & 4.70\textsubscript{$\pm$0.17}  \\
        & MEMIT & 
        1.36\textsubscript{$\pm$0.36} & 1.24\textsubscript{$\pm$0.29} & 0.13\textsubscript{$\pm$0.05} & 4.55\textsubscript{$\pm$0.01} & 75.75\textsubscript{$\pm$0.06} & 70.03\textsubscript{$\pm$0.02} & 40.04\textsubscript{$\pm$0.47} & 73.43\textsubscript{$\pm$0.86} \\
        & PRUNE & 24.99\textsubscript{$\pm$2.43} & 24.45\textsubscript{$\pm$2.34} & 18.02\textsubscript{$\pm$1.69} & 40.57\textsubscript{$\pm$1.61} & 37.24\textsubscript{$\pm$2.20} & 35.95\textsubscript{$\pm$1.59} & 27.91\textsubscript{$\pm$0.17} & 60.35\textsubscript{$\pm$0.89} \\
        & RECT & 79.98\textsubscript{$\pm$1.03} & 74.50\textsubscript{$\pm$0.88} & 42.69\textsubscript{$\pm$0.17} & 72.43\textsubscript{$\pm$0.73} & 75.73\textsubscript{$\pm$0.20} & 68.80\textsubscript{$\pm$0.55} & \textbf{41.52}\textsubscript{$\pm$1.11} & \underline{73.69}\textsubscript{$\pm$0.82} \\
        & AlphaEdit & \underline{93.50}\textsubscript{$\pm$0.18} & \textbf{87.18}\textsubscript{$\pm$0.50} & \underline{42.58}\textsubscript{$\pm$0.29} & \underline{73.01}\textsubscript{$\pm$0.74} & \underline{82.41}\textsubscript{$\pm$1.86} & \underline{73.57}\textsubscript{$\pm$0.53} & 40.08\textsubscript{$\pm$0.48} & 73.56\textsubscript{$\pm$0.62} \\
        & LangEdit & \textbf{93.90}\textsubscript{$\pm$0.04}* & \underline{87.02}\textsubscript{$\pm$0.41} & \textbf{42.64}\textsubscript{$\pm$0.32} & \textbf{74.06}\textsubscript{$\pm$1.33}* & \textbf{83.47}\textsubscript{$\pm$0.91}* & \textbf{74.32}\textsubscript{$\pm$0.19}* & \underline{40.55}\textsubscript{$\pm$0.41} & \textbf{75.70}\textsubscript{$\pm$0.36}* \\
        \bottomrule
        \multirow{8}{*}{\rotatebox{90}{GPT-J-6B}} & Pre-edited & 24.05\textsubscript{$\pm$0.12} & 23.71\textsubscript{$\pm$0.23} & 26.07\textsubscript{$\pm$0.13} & 37.5\textsubscript{$\pm$0.20} & 14.51\textsubscript{$\pm$0.10} & 13.92\textsubscript{$\pm$0.17} & 15.08\textsubscript{$\pm$0.09} & 33.66\textsubscript{$\pm$0.23} \\
        \hline
        & FT & 23.58\textsubscript{$\pm$0.10} & 21.16\textsubscript{$\pm$0.13} & 1.64\textsubscript{$\pm$0.07} & 4.67\textsubscript{$\pm$0.09} & 20.86\textsubscript{$\pm$0.10} & 19.55\textsubscript{$\pm$0.13} & 4.04\textsubscript{$\pm$0.17} & 5.06\textsubscript{$\pm$0.11} \\
        & ROME & 19.66\textsubscript{$\pm$0.10} & 18.37\textsubscript{$\pm$0.38} & 1.36\textsubscript{$\pm$0.10} & 4.93\textsubscript{$\pm$0.15} &  14.41\textsubscript{$\pm$0.37} & 13.08\textsubscript{$\pm$0.30} & 0.78\textsubscript{$\pm$0.15} & 6.52\textsubscript{$\pm$1.78} \\
        & MEMIT & 48.25\textsubscript{$\pm$4.25} & 46.07\textsubscript{$\pm$4.14} & 22.63\textsubscript{$\pm$1.08} & 36.85\textsubscript{$\pm$0.76} & 44.98\textsubscript{$\pm$0.22} & 41.75\textsubscript{$\pm$0.36} & 14.47\textsubscript{$\pm$0.16} & 31.53\textsubscript{$\pm$0.76} \\
        & PRUNE  & 3.10\textsubscript{$\pm$0.83} & 2.93\textsubscript{$\pm$0.86} & 2.40\textsubscript{$\pm$0.62} & 8.26\textsubscript{$\pm$1.88} & 2.41\textsubscript{$\pm$0.38} & 2.42\textsubscript{$\pm$0.44} & 2.37\textsubscript{$\pm$0.39} & 5.77\textsubscript{$\pm$3.44} \\
        & RECT & 71.10\textsubscript{$\pm$1.78} & 67.05\textsubscript{$\pm$1.99} & 26.26\textsubscript{36} & \underline{37.17}\textsubscript{$\pm$0.95} & 48.30\textsubscript{$\pm$0.30} & 43.33\textsubscript{$\pm$0.56} & 14.48\textsubscript{$\pm$0.17} & 32.67\textsubscript{$\pm$1.29} \\
        & AlphaEdit & \underline{83.59}\textsubscript{$\pm$0.26} & \underline{78.34}\textsubscript{$\pm$0.05} & \underline{26.55}\textsubscript{$\pm$0.33} & 36.74\textsubscript{$\pm$1.19} & \underline{54.36}\textsubscript{$\pm$0.11} & \underline{47.52}\textsubscript{$\pm$0.15} & \underline{15.13}\textsubscript{$\pm$0.20} & \underline{33.28}\textsubscript{$\pm$1.21} \\
        & LangEdit & \textbf{84.27}\textsubscript{$\pm$0.27}* & \textbf{79.74}\textsubscript{$\pm$0.36}* & \textbf{27.23}\textsubscript{$\pm$0.04}* & \textbf{38.59}\textsubscript{$\pm$1.35}* & \textbf{54.86}\textsubscript{$\pm$0.21}* & \textbf{48.40}\textsubscript{$\pm$0.44}* & \textbf{15.31}\textsubscript{$\pm$0.05} & \textbf{35.75}\textsubscript{$\pm$0.94}* \\
        \bottomrule
    \end{tabular}
    }
    \caption{We assess the performance of various model editing methods using three LLMs (GPT-J-6B, Llama3-8B, and Qwen2.5-7B) on the mzsre and bzsre datasets. The best results are highlighted in \textbf{bold}, while the second-best results are underlined. Statistical significance (*) is determined using a paired t-test with p=0.05. XTREME$\ddag$ represents  average F1 Scores on XTREME tasks after training on the mzsre dataset; XTREME$\dag$ denotes the average F1 Scores after training on the bzsre dataset.  All baselines are adapted for multilingual sequential knowledge editing.}
    \label{tab:full_model_editing}
\end{table*}

\subsubsection{Implementation Details}
We perform multilingual sequential knowledge editing with 100 samples per edit time step $t$. We conducted all experiments on a single A100 GPU (80GB) and repeated three times with different seeds to ensure reliability. Results are reported as the mean and standard deviation.
We evaluate the computational cost of different editing methods in the Appendix~\ref{sec:computation}.
Following~\cite{mengmass,fang2024alphaedit}, we configure each model as follows:
(1) For the GPT-J-6B model, we target critical layers [3, 4, 5, 6, 7, 8] for editing inspired by monolingual editing. When computing the hidden representations of a critical layer, we perform 25 optimization steps with a learning rate of 0.5.
(2) For the Llama3-8B model, we target critical layers [4, 5, 6, 7, 8] for editing. 
When computing the hidden representations of a critical layer, we perform 25 steps with a learning rate of 0.1.
(3) For the Qwen2.5-7B model, we target critical layers [4, 5, 6, 7, 8] for editing. When computing the hidden representations of a critical layer, we perform 25 steps with a learning rate of 0.1.
The selection of critical layers is guided by the causal tracing technique, with detailed results provided in Appendix~\ref{sec:analysis_fact}. 
For the package versions, we report them in the Appendix~\ref{sec:package}.

\subsubsection{Baselines}

This work establishes the first benchmark for multilingual sequential knowledge editing. 
As no existing methods specifically address this emerging challenge, we establish baseline performance by adapting state-of-the-art approaches from monolingual sequential editing research. We faithfully re-implemented these models based on their original codebases.

\noindent \textbf{FT} (Fine-Tuning) updates a subset of the model parameters via gradient descent using the training examples from the multilingual knowledge editing dataset. While effective for adapting models to new distributions, FT risks catastrophic forgetting of pre-trained knowledge and demands substantial computational resources.

\noindent \textbf{ROME} (Rank-One Model Editing)~\cite{meng2022locating} localizes factual associations to middle-layer feed-forward modules and updates their weights through rank-one adjustments. By identifying critical neuron activations tied to specific knowledge, ROME demonstrates that direct manipulation of feed-forward layers can edit factual predictions without requiring full retraining.

\noindent \textbf{MEMIT} (Mass-Editing Memory in Transformer)~\cite{mengmass} extends ROME by propagating edits across multiple layers. It identifies critical neuron activations in parallelized weight matrices during knowledge editing, enabling the insertion of thousands of new associations.

\noindent \textbf{PRUNE} (Perturbation
Restraint on Upper bouNd for Editing)~\cite{ma2024perturbation} mitigates performance degradation during sequential editing by optimizing the condition number~\cite{smith1967condition} of edited weights, which limits catastrophic interference with unrelated knowledge as the number of edits accumulates.

\noindent \textbf{RECT} (RElative Change in weighT)~\cite{gu-etal-2024-model} mitigates the side effects of editing on general reasoning abilities by controlling weight updates.  
Specifically, the top-k\% parameters that change the most according to relative changes in parameters are considered as the principal editing information and their obtained values are kept, while the remaining parameters are kept unchanged. 

\noindent \textbf{AlphaEdit} \cite{fang2024alphaedit} is designed for monolingual knowledge editing and decouples knowledge updates from preservation objectives by null-space projection. Parameter perturbations in AlphaEdit are projected onto a static null space of key matrices, whereas LangEdit leverages a dynamic null space, where the projection of the key matrices varies at each update step $t$.
The difference between the original AlphaEdit and our method is two-fold. The first difference is that the original AlphaEdit and our method leverage the null-space projection to resolve a different task. The original AlphaEdit focuses on monolingual sequential knowledge editing and our work solves the task of multilingual sequential knowledge editing. The second difference is the usage of null space projection. Parameter perturbations in AlphaEdit are projected onto a static null space of key matrices, while LangEdit leverages a dynamic null space, where the projection of the key matrices is different for any update step. 

Baseline models (e.g., MEMIT, AlphaEdit) are designed to support  batch editing (i.e., 100 samples per time-step). As ROME does not support batch-editing, we run ROME 100 times iteratively to adapt the multilingual sequential knowledge editing task. 

Moreover, we propose several baseline methods derived from our model architecture, with comprehensive comparison results provided in Appendix~\ref{sec:var_lang}. We also evaluate the pre-edited model — the original LLMs without any editing.

\begin{figure*}[ht]
\centering
\subfloat[\scriptsize{MLQA}]{\includegraphics[width=0.23\textwidth]{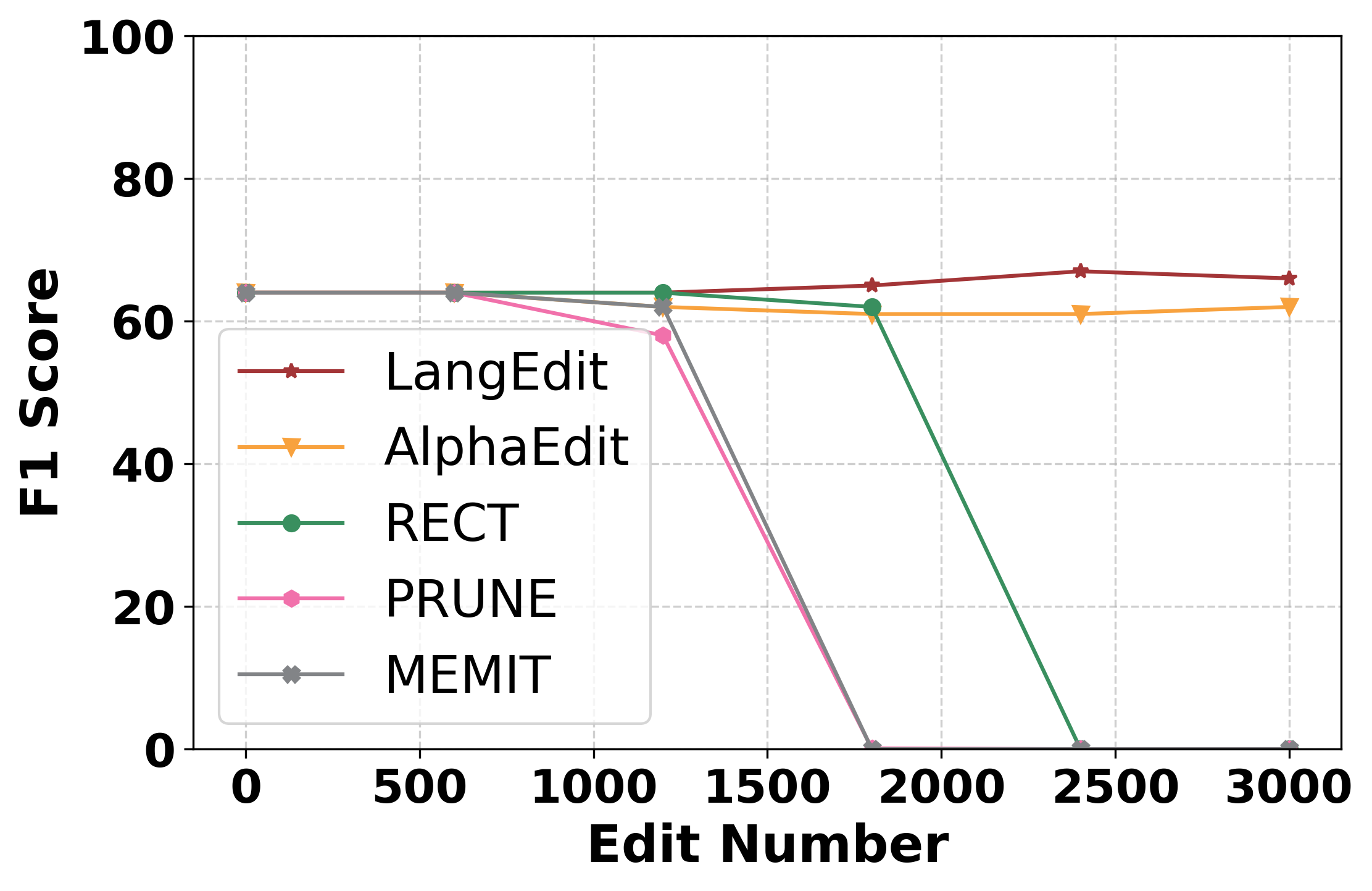}
\label{fig:exp_line_downstream_mlqa}
}
\hfil
\subfloat[\scriptsize{Wikiann}]{\includegraphics[width=0.23\textwidth]{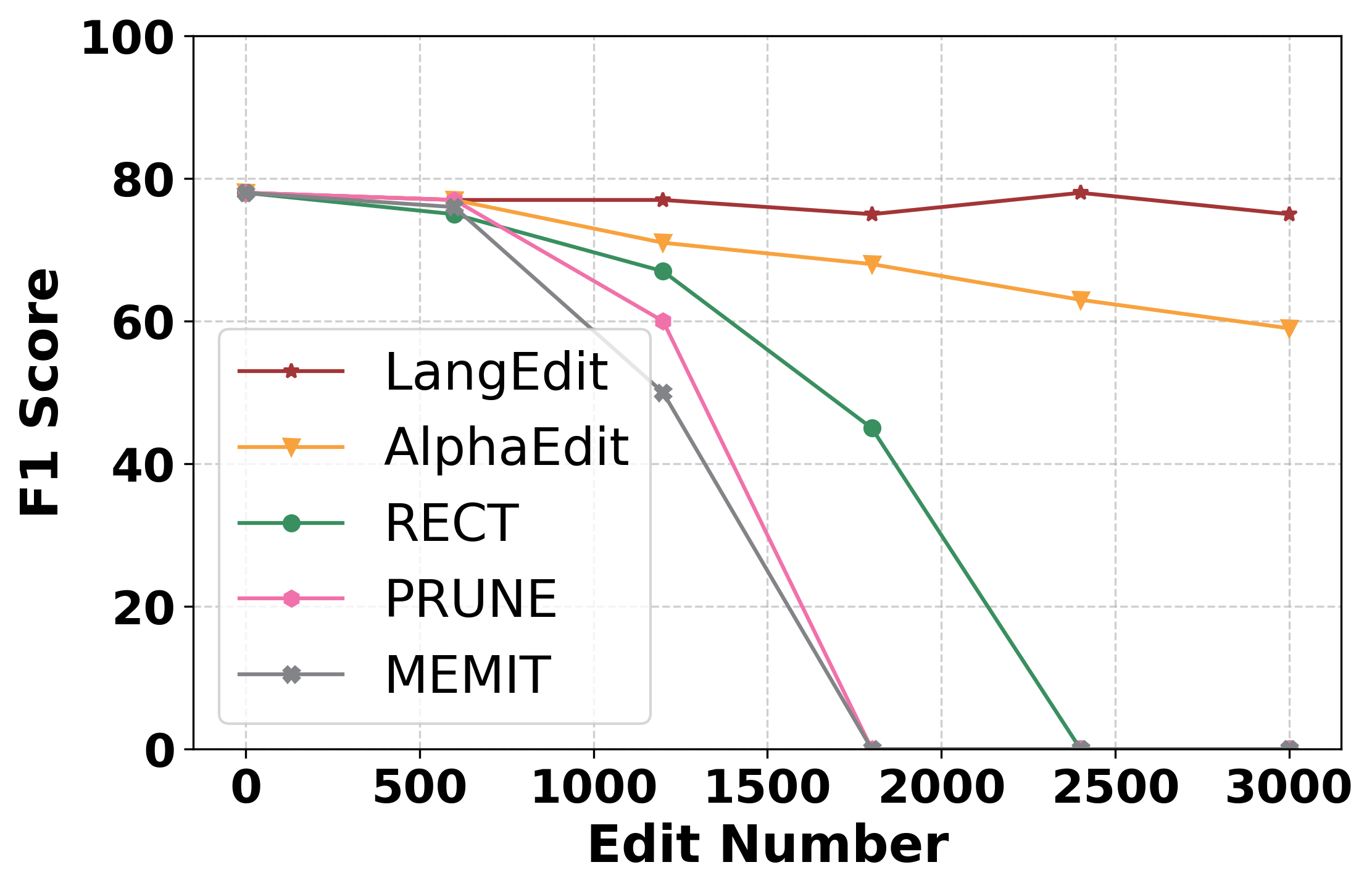}
\label{fig:exp_line_downstream_wikiann}
}
\hfil
\subfloat[\scriptsize{PAWSX}]{\includegraphics[width=0.23\textwidth]{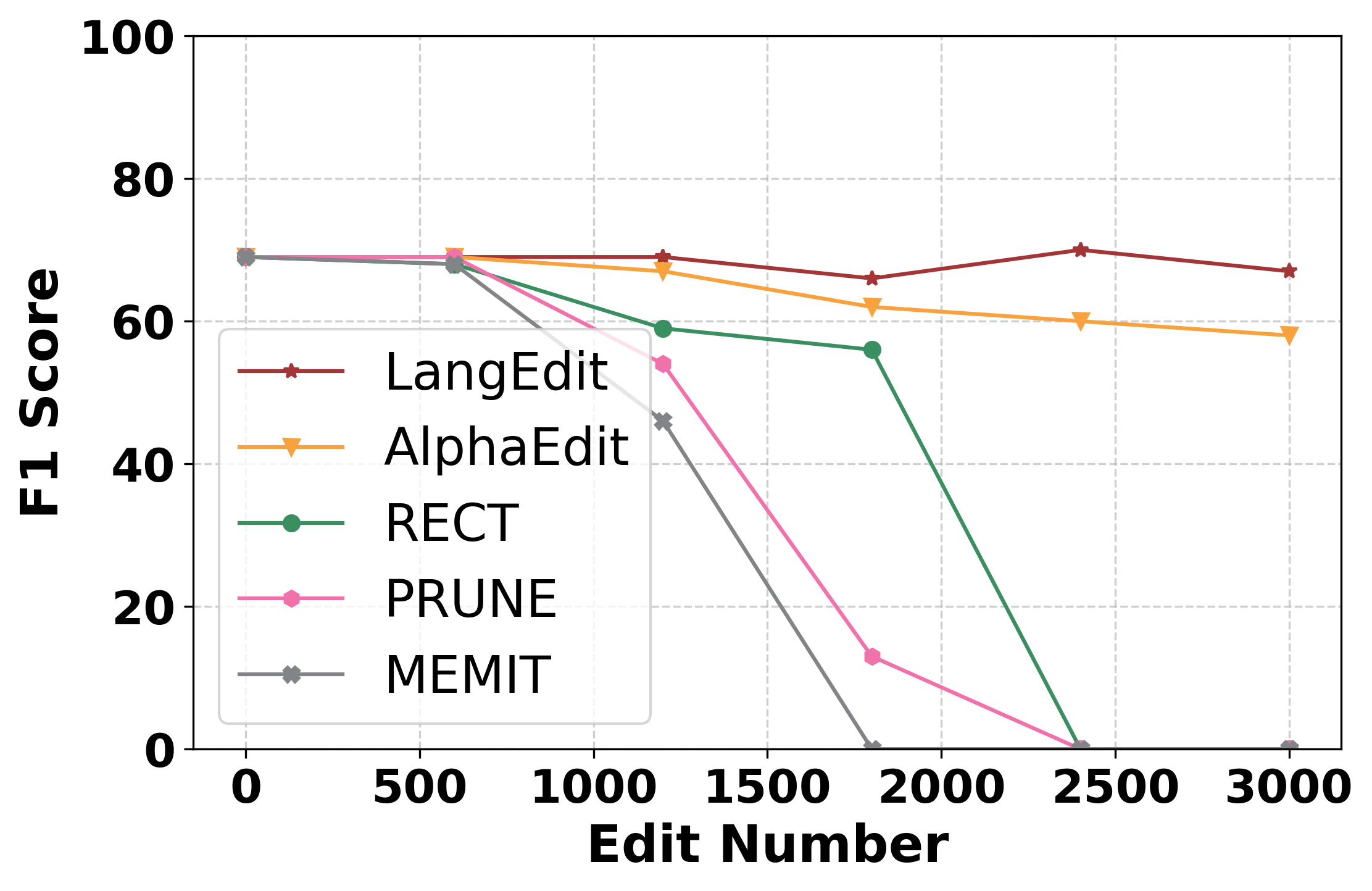}
\label{fig:exp_line_downstream_pawsx}
}
\hfil
\subfloat[\scriptsize{XNLI}]{\includegraphics[width=0.23\textwidth]{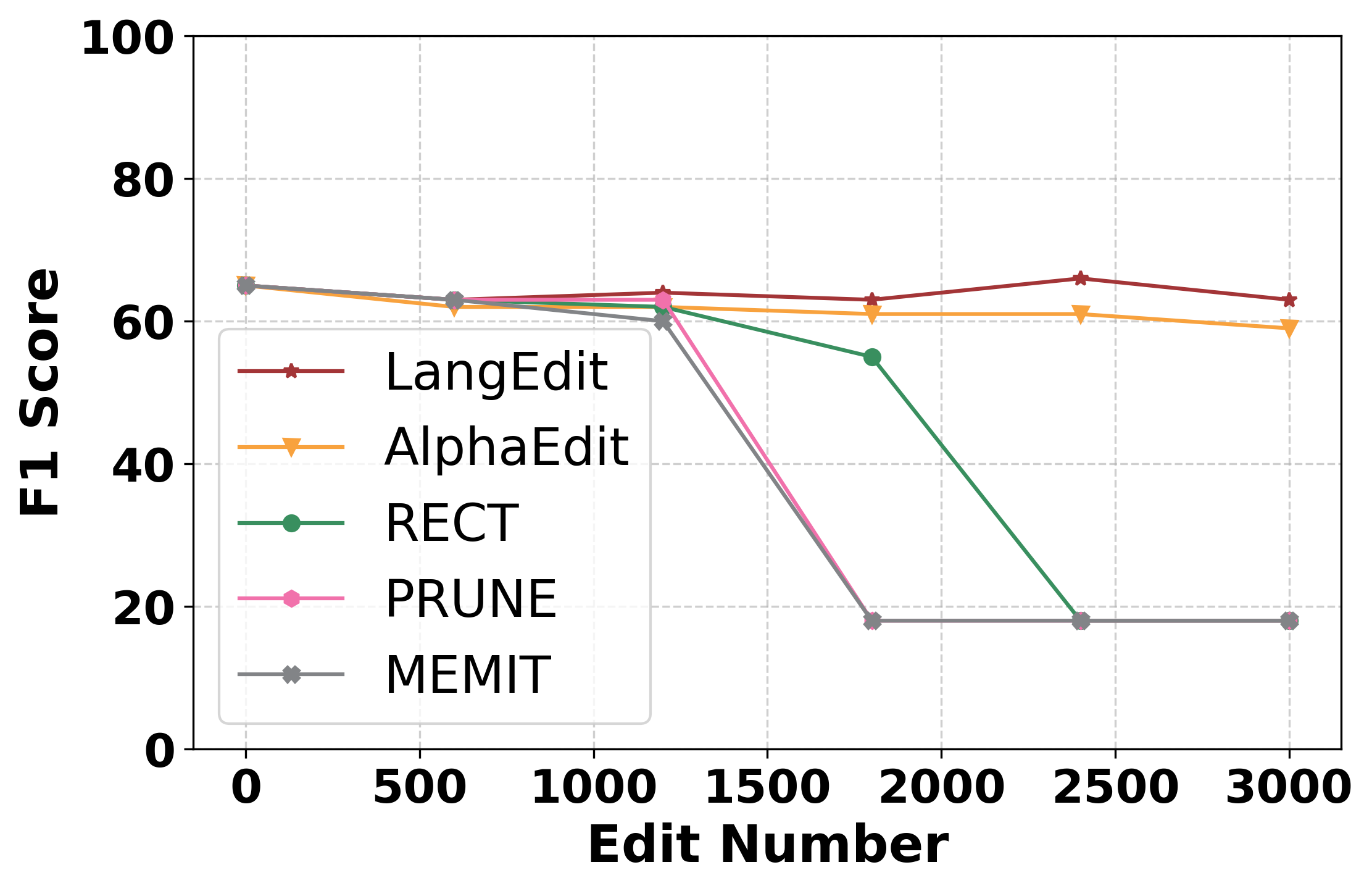}
\label{fig:exp_line_downstream_xnli}
}
\caption{F1 Scores of the edited Llama3-8B on XTREME benchmark evaluating multilingual generalization.}
\label{fig:exp_line_downstream}
\end{figure*}

\begin{figure*}[htbp]
\centering
\subfloat[\scriptsize{EN (mzsre))}]{\includegraphics[width=0.2\textwidth]{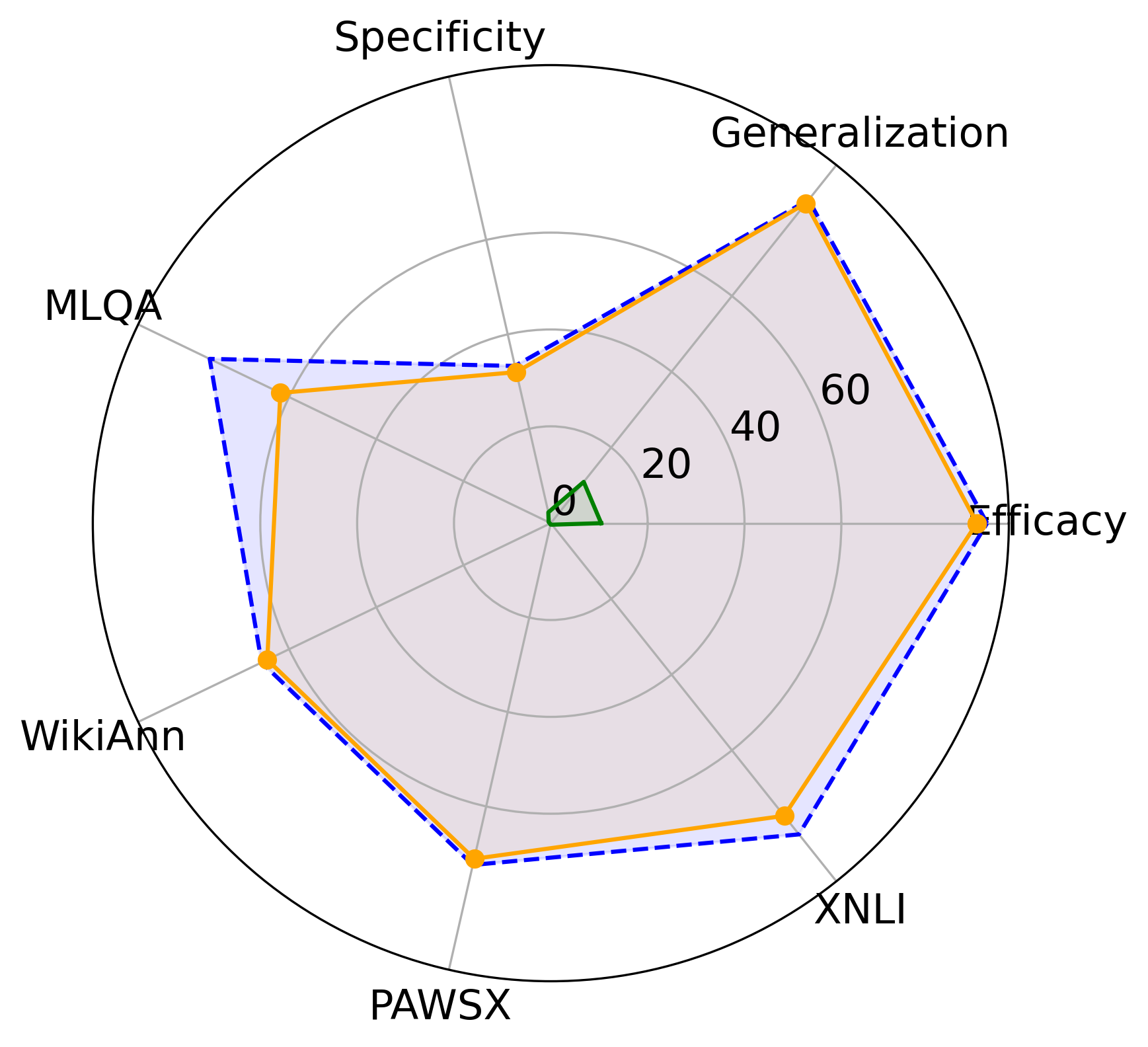}
\label{fig:exp_radar_full_mzsre_en_2400}
}
\hfil
\subfloat[\scriptsize{DE (mzsre)}]{\includegraphics[width=0.2\textwidth]{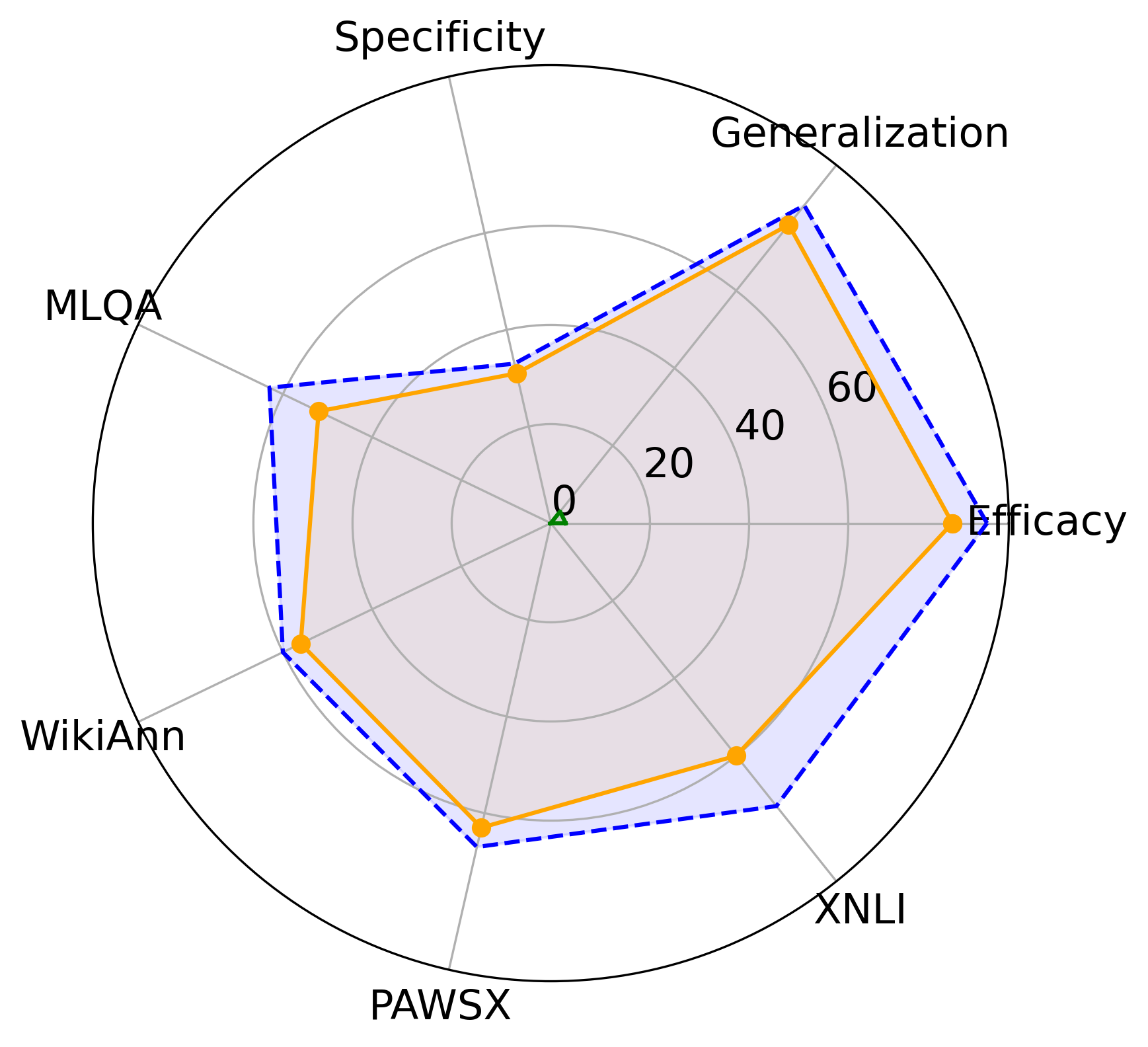}
\label{fig:exp_radar_full_mzsre_de_2400}
}
\hfil
\subfloat[\scriptsize{NL (mzsre)}]{\includegraphics[width=0.2\textwidth]{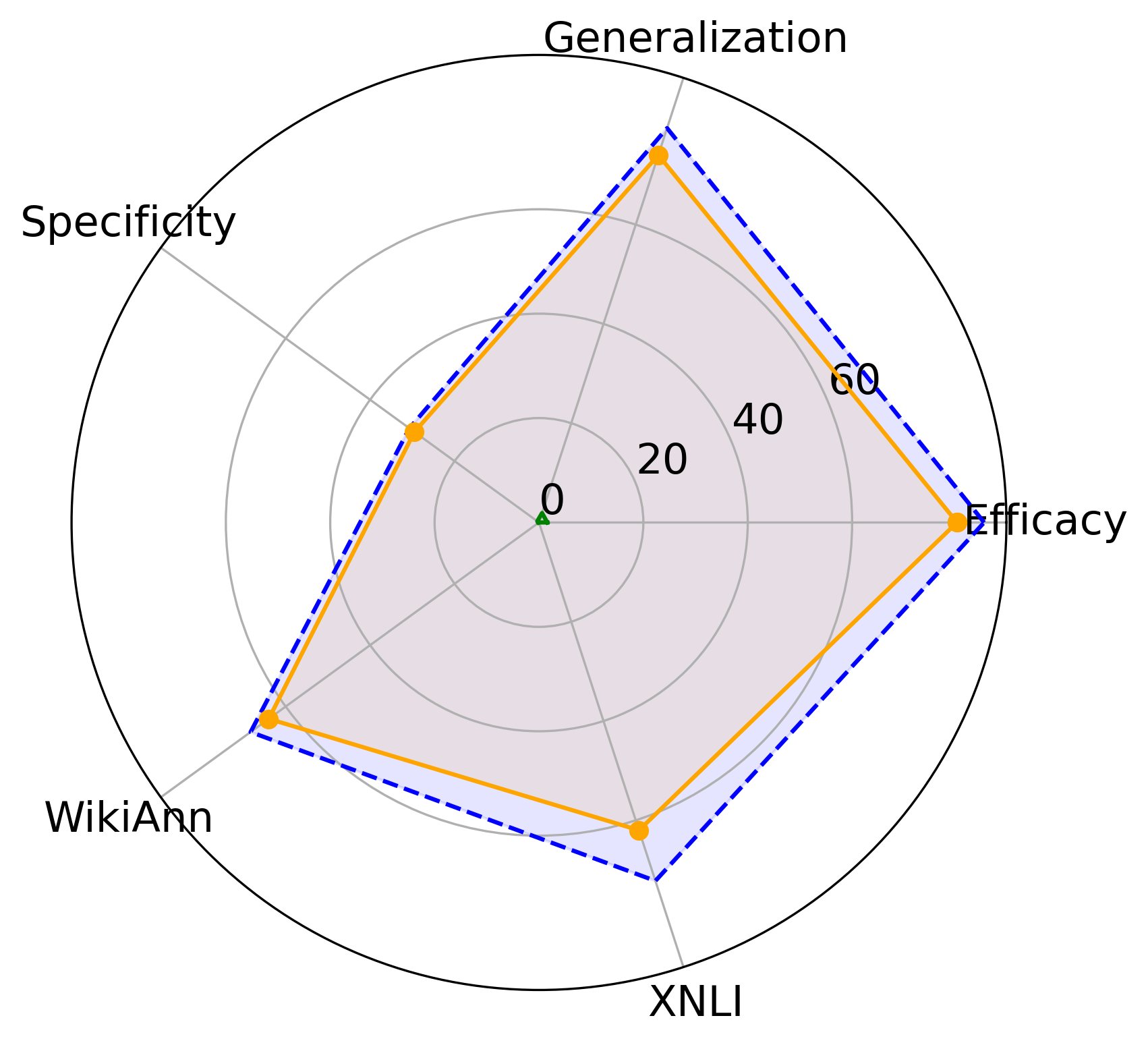}
\label{fig:exp_radar_full_mzsre_nl_2400}
}
\hfil
\subfloat[\scriptsize{ES (mzsre)}]{\includegraphics[width=0.2\textwidth]{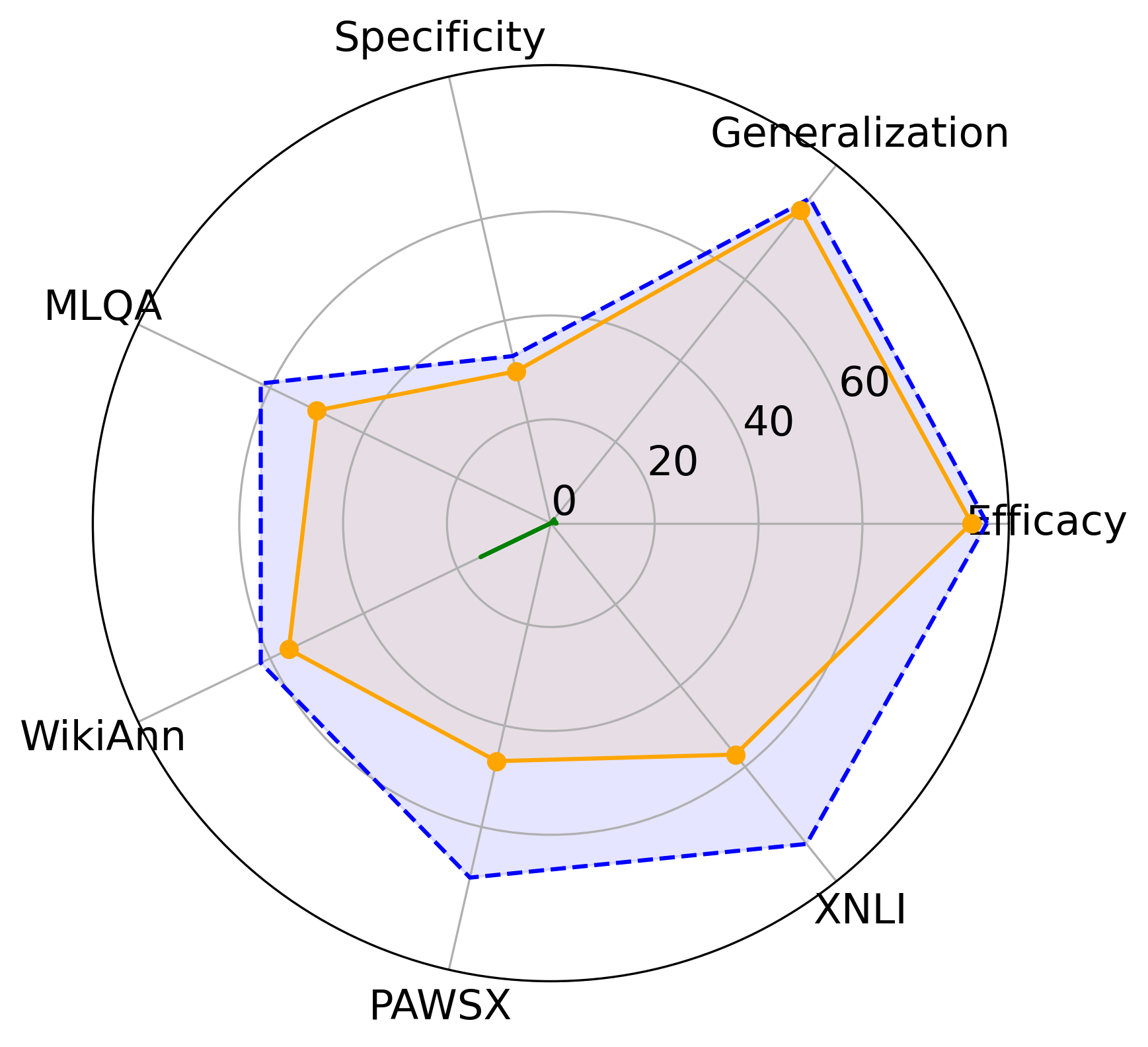}
\label{fig:exp_radar_full_mzsre_es_2400}
}
\hfil
\subfloat[\scriptsize{FR (mzsre)}]{\includegraphics[width=0.2\textwidth]{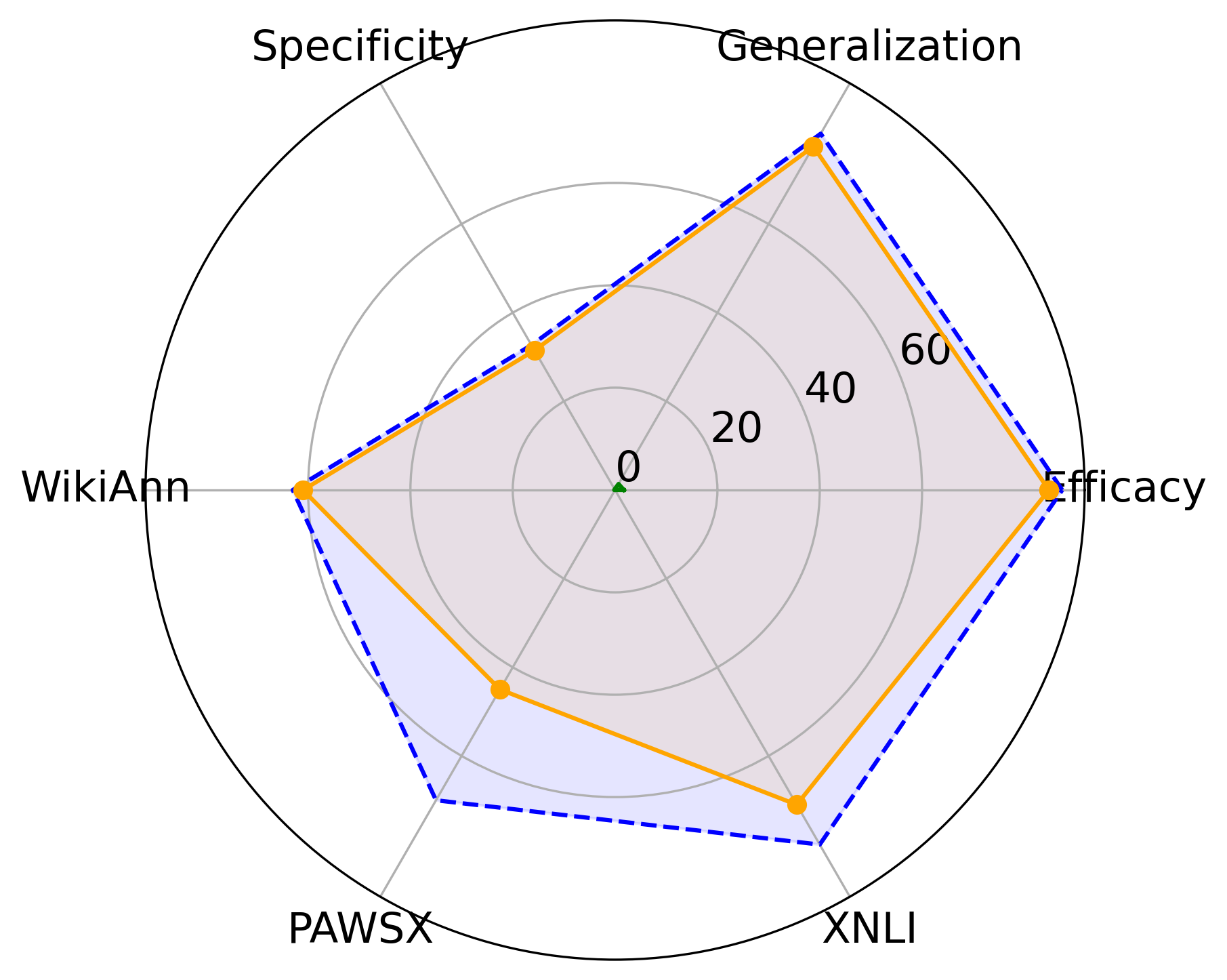}
\label{fig:exp_radar_full_mzsre_fr_2400}
}
\hfil
\subfloat[\scriptsize{ZH (mzsre)}]{\includegraphics[width=0.2\textwidth]{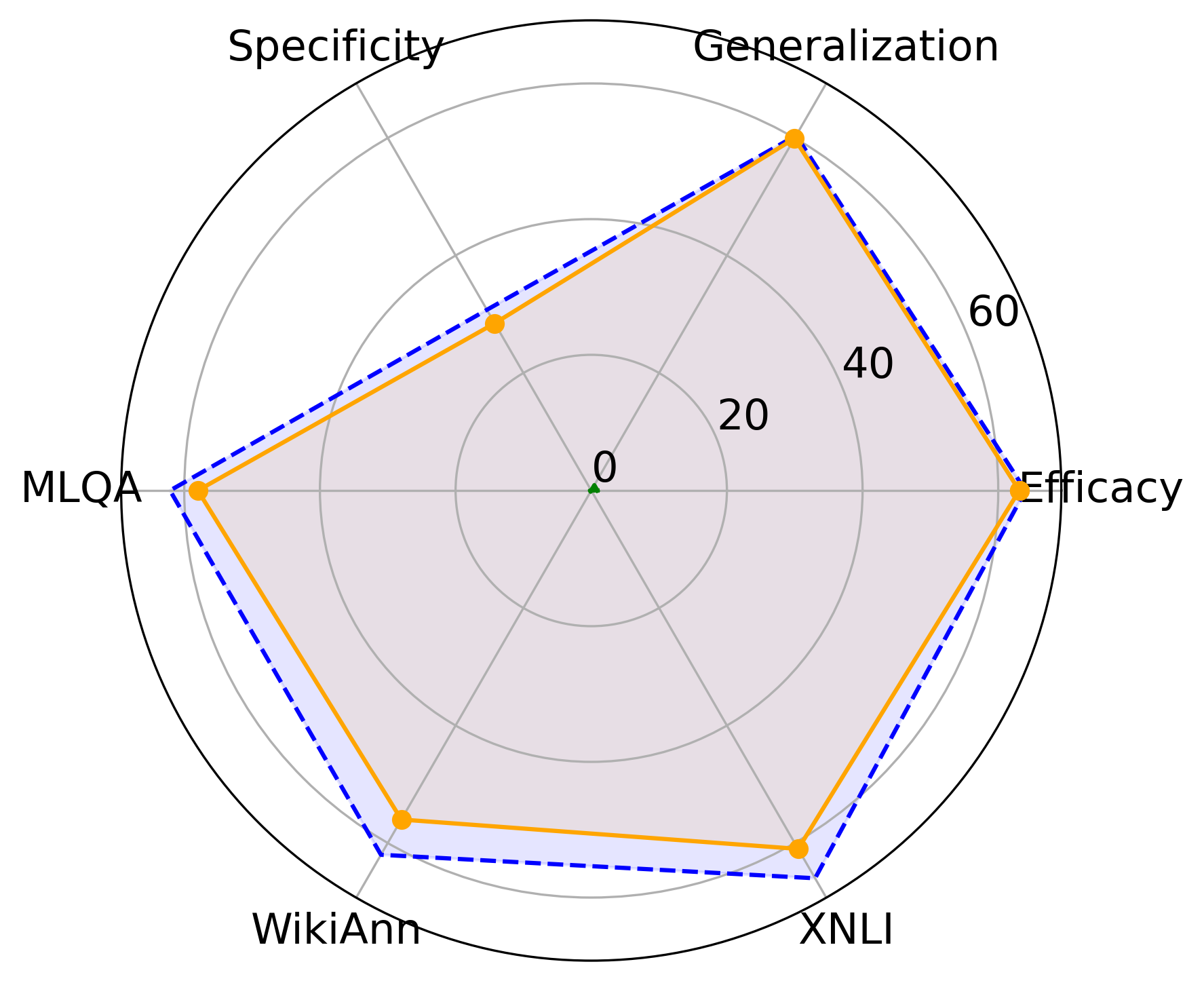}
\label{fig:exp_radar_full_mzsre_zh_2400}
}
\hfil
\subfloat[\scriptsize{EN (bzsre)}]{\includegraphics[width=0.2\textwidth]{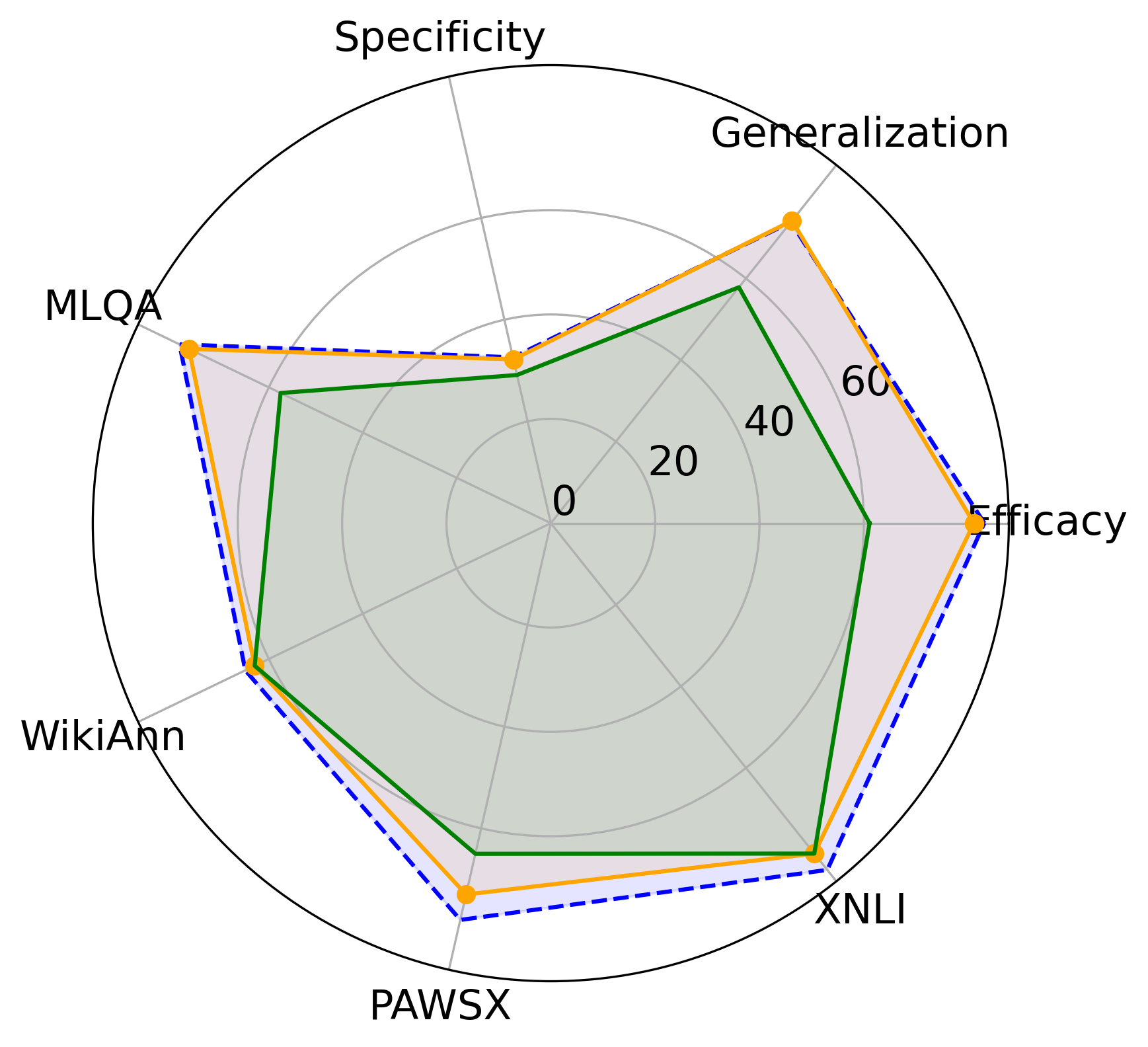}
\label{fig:exp_radar_full_bizsre_en_1600}
}
\hfil
\subfloat[\scriptsize{ZH (bzsre)}]{\includegraphics[width=0.2\textwidth]{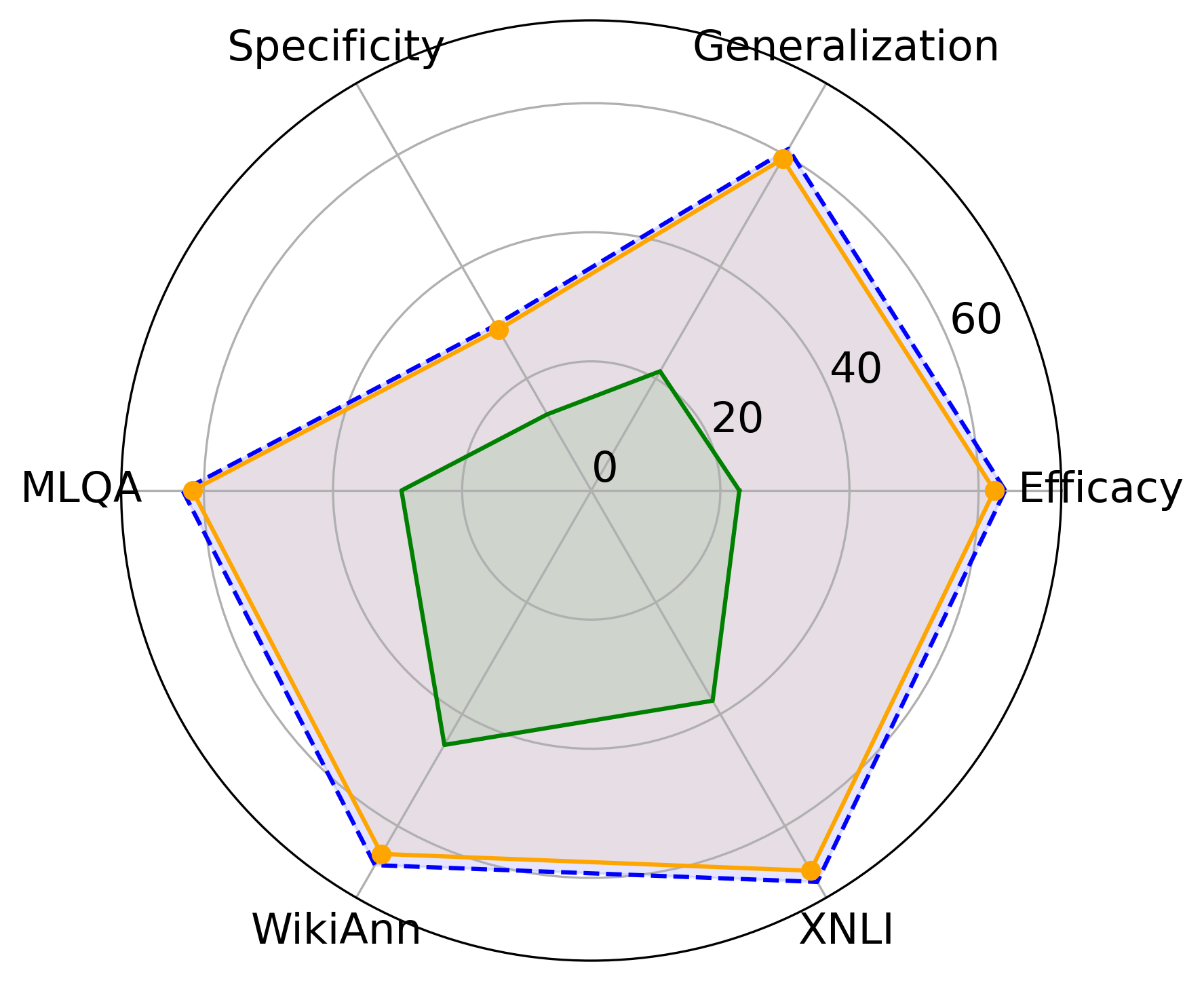}
\label{fig:exp_radar_full_bizsre_zh_1600}
}
\caption{Radar chart of editing performance and multilingual downstream task performance when editing Llama3-8B with different languages. We use \dashedline{}, \orangeline{} and \greenline{} to represent LangEdit, AlphaEdit and RECT, respectively.}
\label{fig:radar_llama}
\end{figure*}

\subsection{Experimental Results}

Table~\ref{tab:full_model_editing} demonstrates that LangEdit significantly outperforms the state-of-the-art model AlphaEdit in terms of Efficacy Score across three model architectures (Llama3-8B, Qwen2.5-7B, GPT-J-6B) and two knowledge editing datasets (mzsre, bzsre). 
When using Llama3-8B as the backbone, LangEdit achieves average improvements of +2.20 and +1.30 percentage points~\footnote{All increases or decreases are given in percentage points.} on the mzsre and bzsre datasets, respectively. 
Similar enhancements are observed with Qwen2.5-7B (+0.40 on mzsre, +1.06 on bzsre) and GPT-J-6B (+0.68 on mzsre, +0.60 on bzsre). 
These results underscore the robustness of our method across diverse architectures and its effectiveness in knowledge injection for LLMs.

Furthermore, Table~\ref{tab:full_model_editing} reveals that LangEdit surpasses the best baseline (AlphaEdit in a multilingual setting) by an average margin of +2.85 F1 Score and +2.17 F1 Score on XTREME benchmark when evaluated across three backbone architectures (GPT-J-6B, Qwen2.5-7B, Llama3-8B) after training on mzsre and bzsre datasets, respectively. 
The averaged F1 Score reflects the performance across four tasks designed to assess multilingual generalization. Notably, the improvement is most pronounced for Llama3-8B, which exhibits a +5.65 F1 Score gain on the XTREME benchmark after training on the mzsre dataset. This highlights LangEdit's ability to preserve and enhance multilingual generalization across diverse model architectures during sequential knowledge editing.

Our findings indicate that multilingual knowledge editing significantly improves multilingual generalization capabilities. For GPT-J-6B and Qwen2.5-7B, LangEdit consistently outperforms their pre-edit counterparts on both datasets. While Llama3-8B shows a 3.06 F1 Score decrease on XTREME benchmark after training on the mzsre dataset, it achieves a 3.76 F1 Score improvement on the XTREME benchmark after training on the bzsre dataset. We hypothesize that this discrepancy arises because the injected multilingual knowledge benefits languages structurally aligned with the editing corpus. This observation aligns with prior work~\cite{chua2024crosslingual}, which suggests that even minimal multilingual exposure can enhance generalization ability.

Figure~\ref{fig:exp_line_downstream} illustrates editing performance and multilingual generalization performance trends under varying edits. A key pattern is:
LangEdit consistently outperforms baselines across all edit scales in tasks evaluating multilingual generalization.

\begin{figure*}[htbp]
\centering
\subfloat[\scriptsize{EN (mzsre))}]{\includegraphics[width=0.1\textwidth]{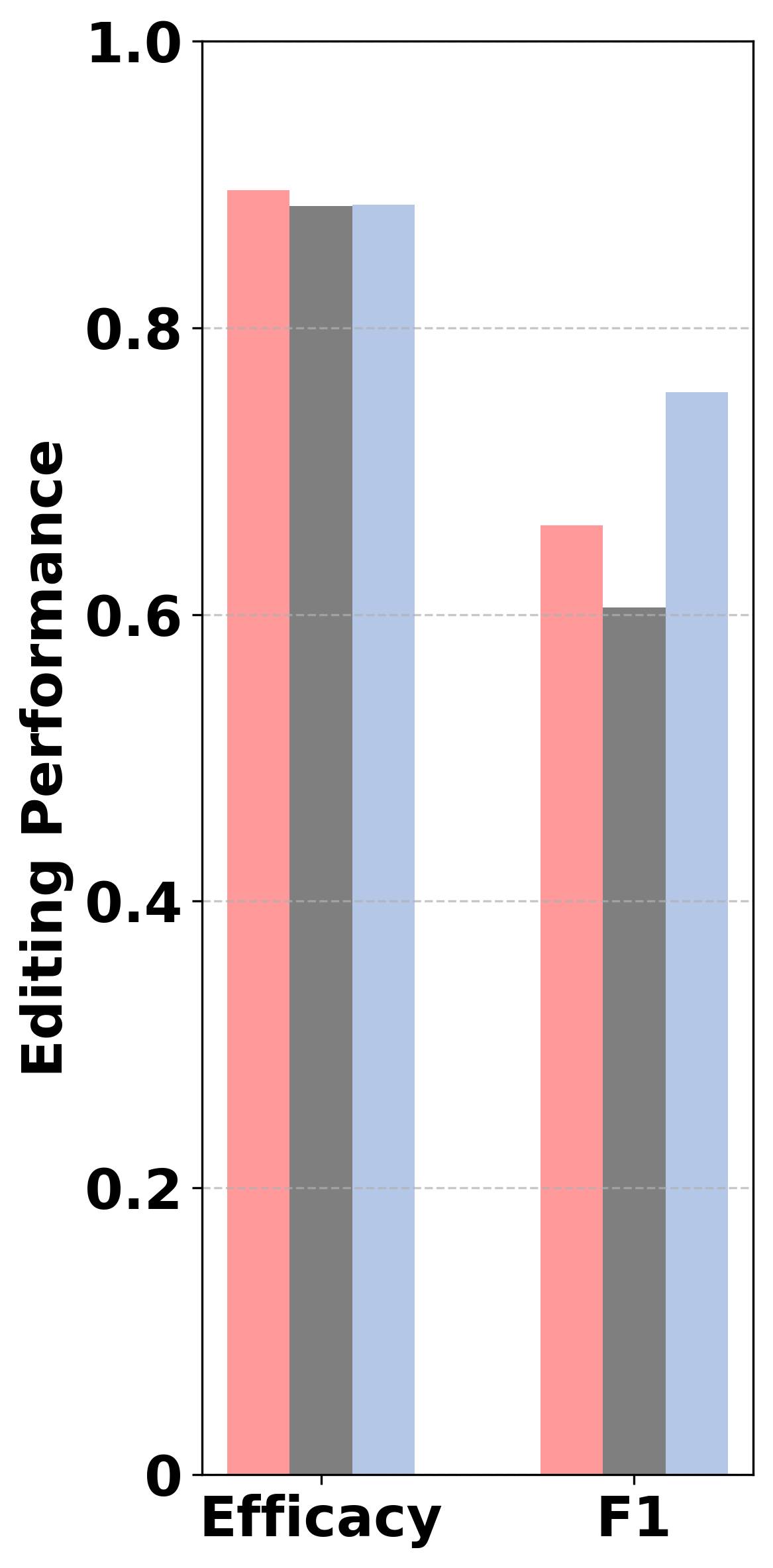}
\label{fig:r1_n}
}
\hfil
\subfloat[\scriptsize{DE (mzsre)}]{\includegraphics[width=0.1\textwidth]{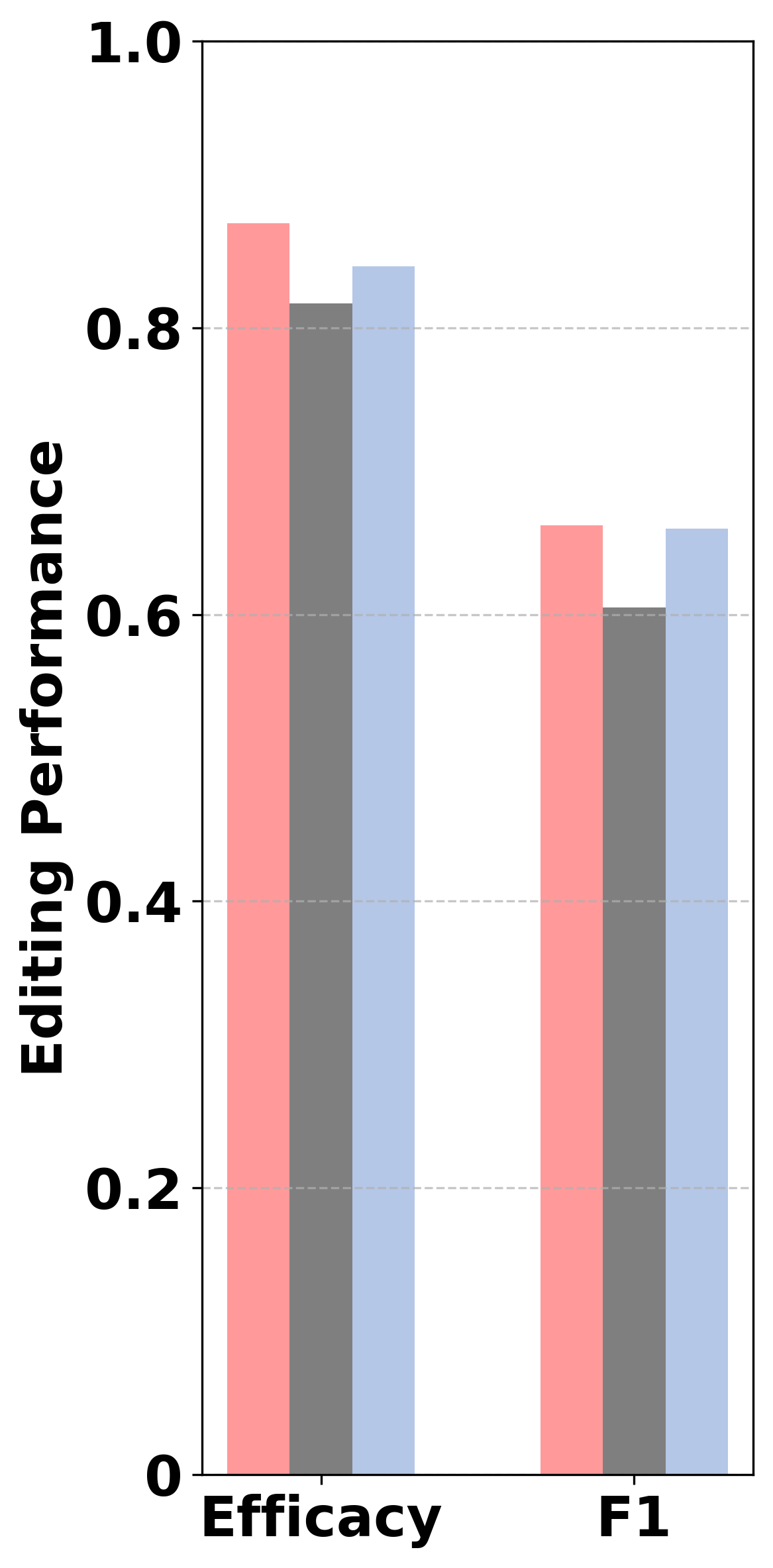}
\label{fig:r2_n}
}
\hfil
\subfloat[\scriptsize{NL (mzsre)}]{\includegraphics[width=0.1\textwidth]{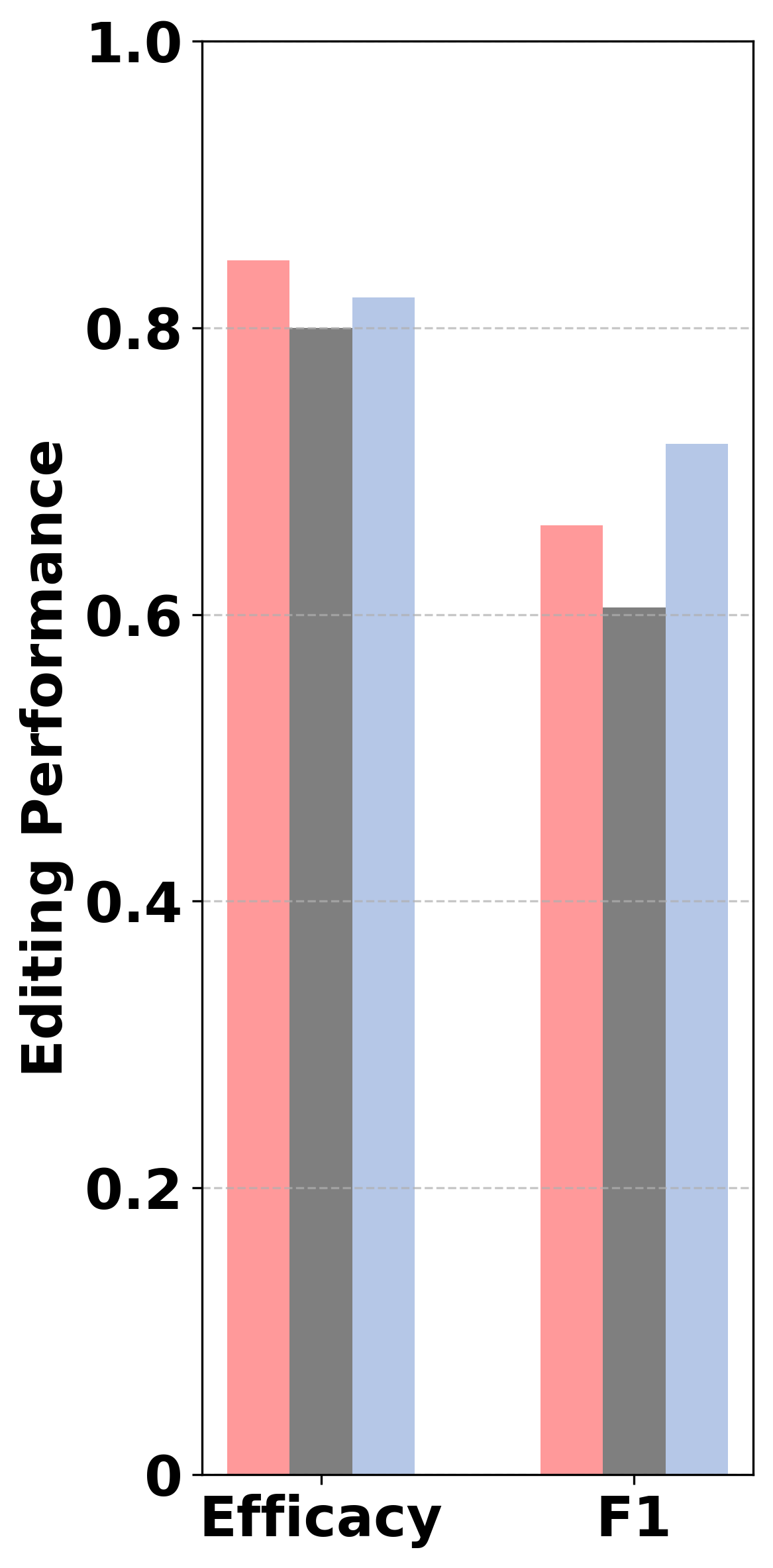}
\label{fig:b4_nl}
}
\hfil
\subfloat[\scriptsize{ES (mzsre)}]{\includegraphics[width=0.1\textwidth]{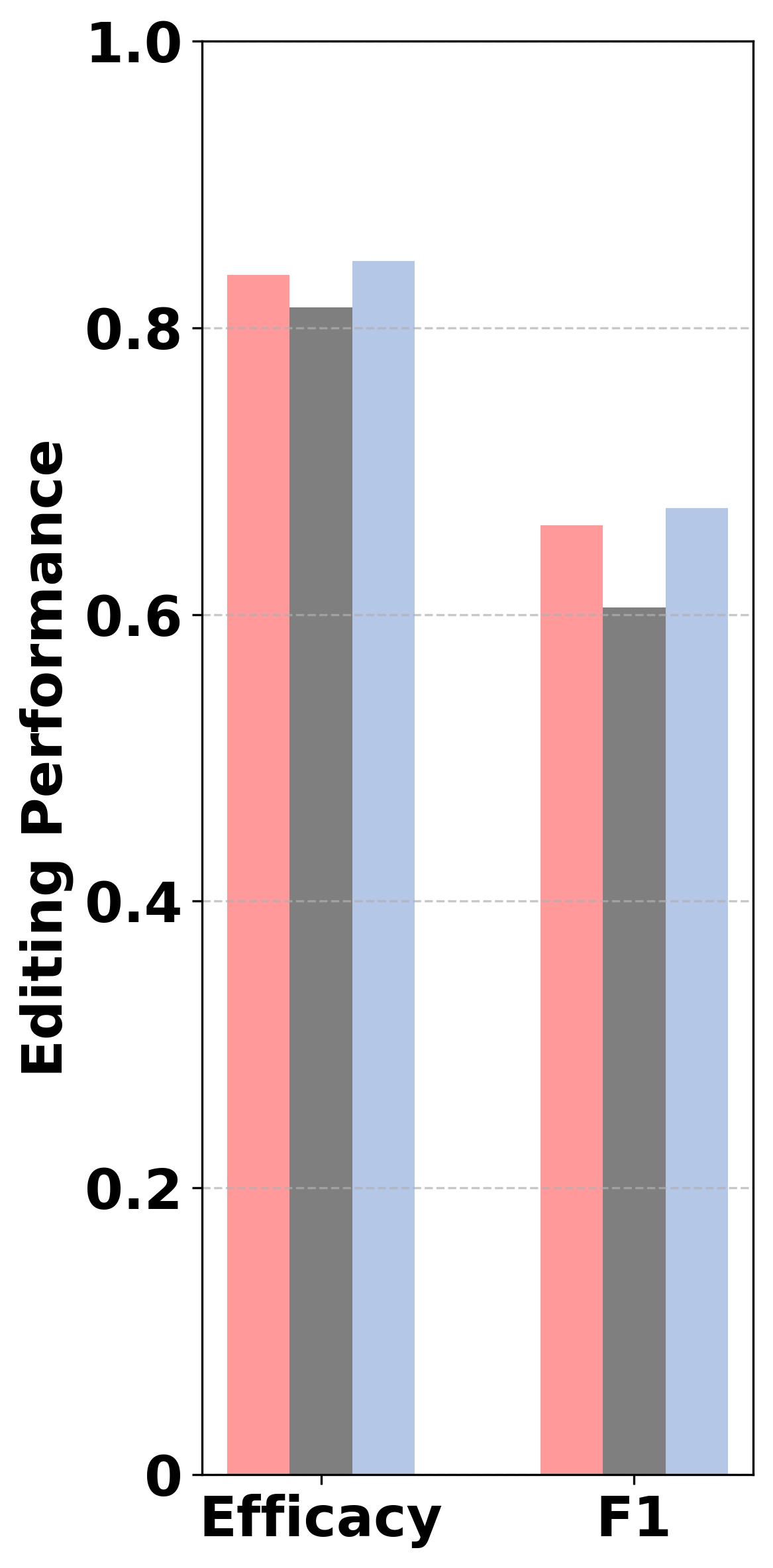}
\label{fig:rl_es}
}
\hfil
\subfloat[\scriptsize{FR (mzsre)}]{\includegraphics[width=0.1\textwidth]{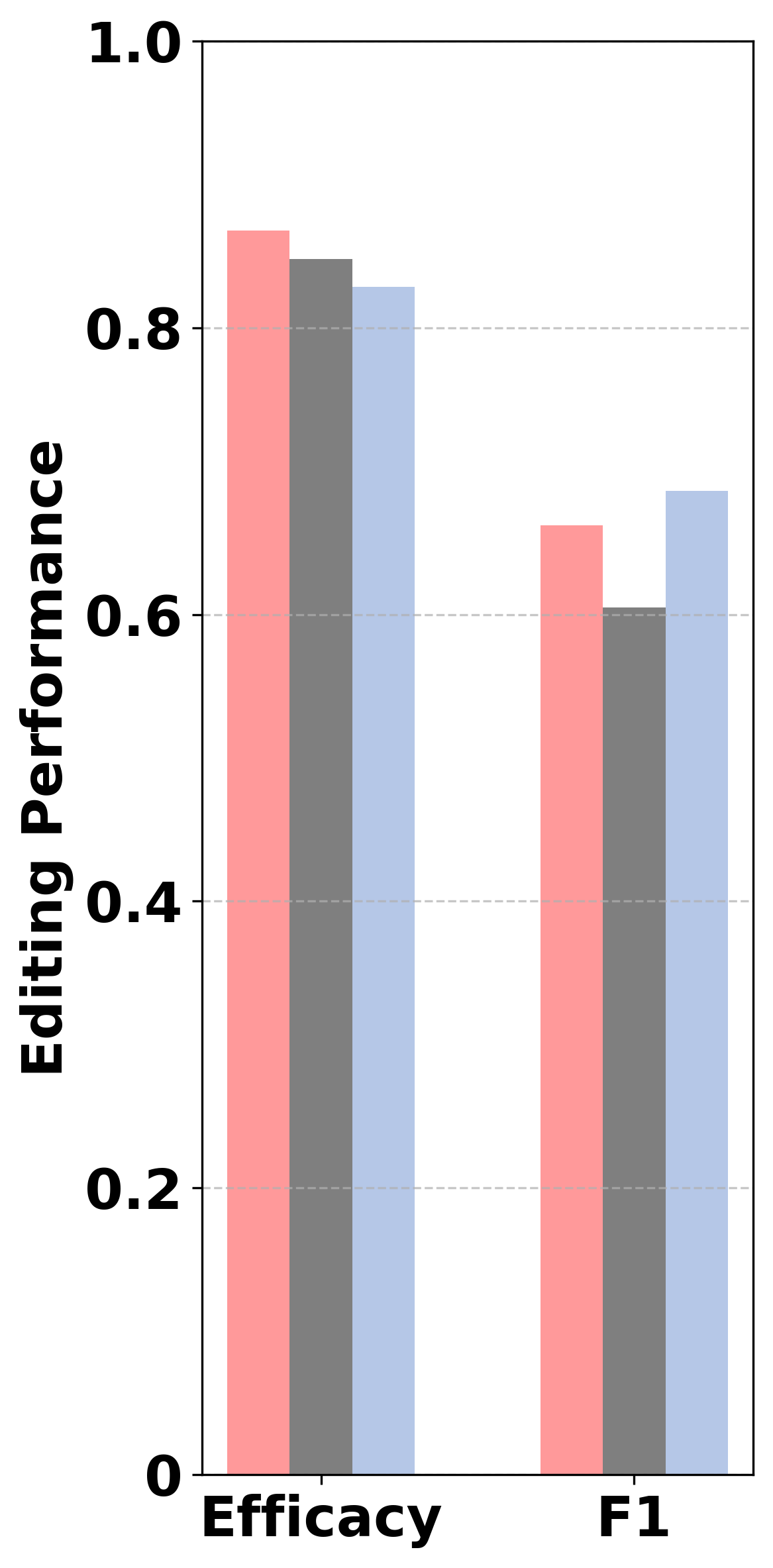}
\label{fig:b4_fr}
}
\hfil
\subfloat[\scriptsize{ZH (mzsre)}]{\includegraphics[width=0.1\textwidth]{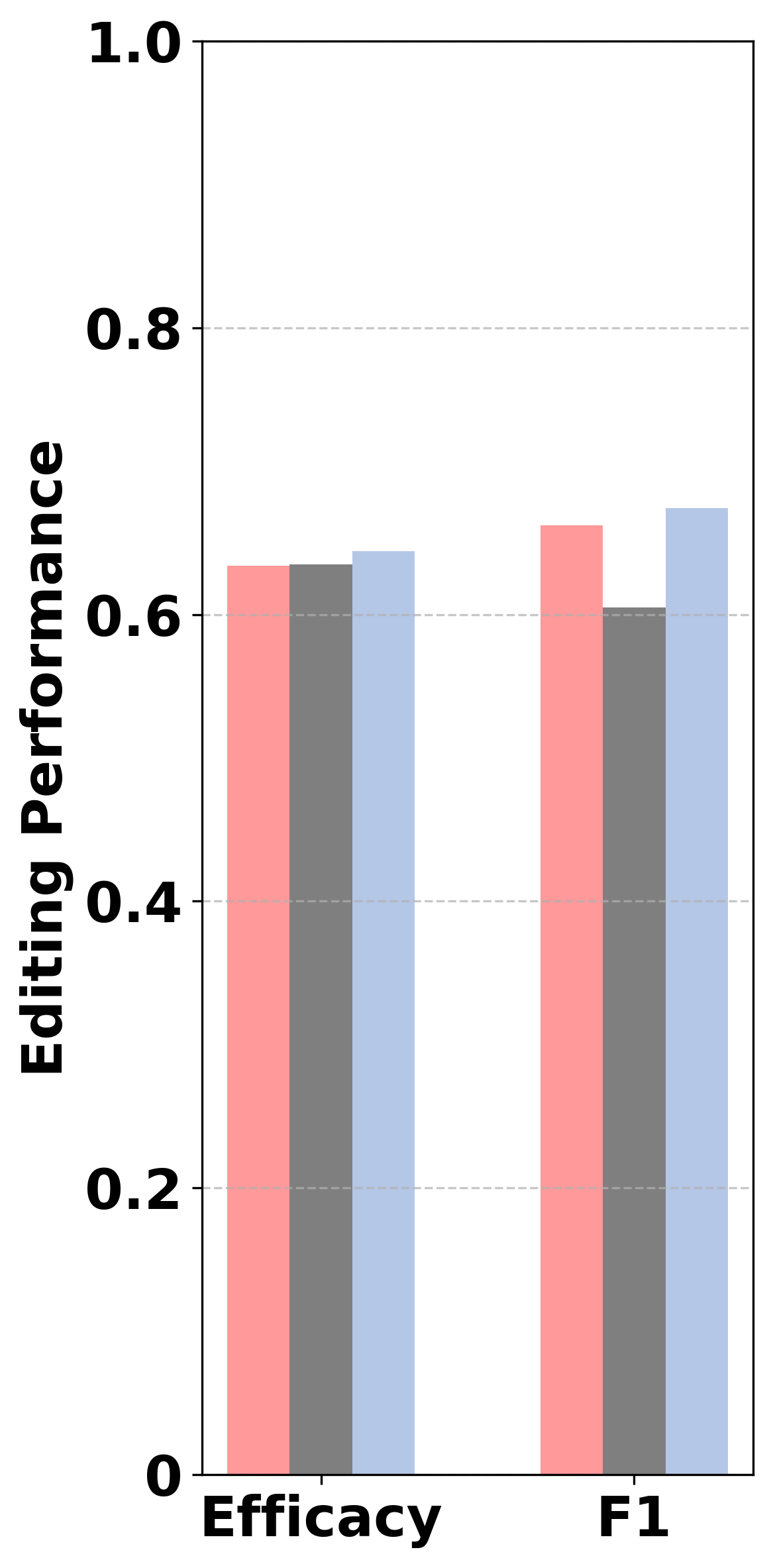}
\label{fig:b4_zh}
}
\hfil
\subfloat[\scriptsize{EN (bzsre)}]{\includegraphics[width=0.1\textwidth]{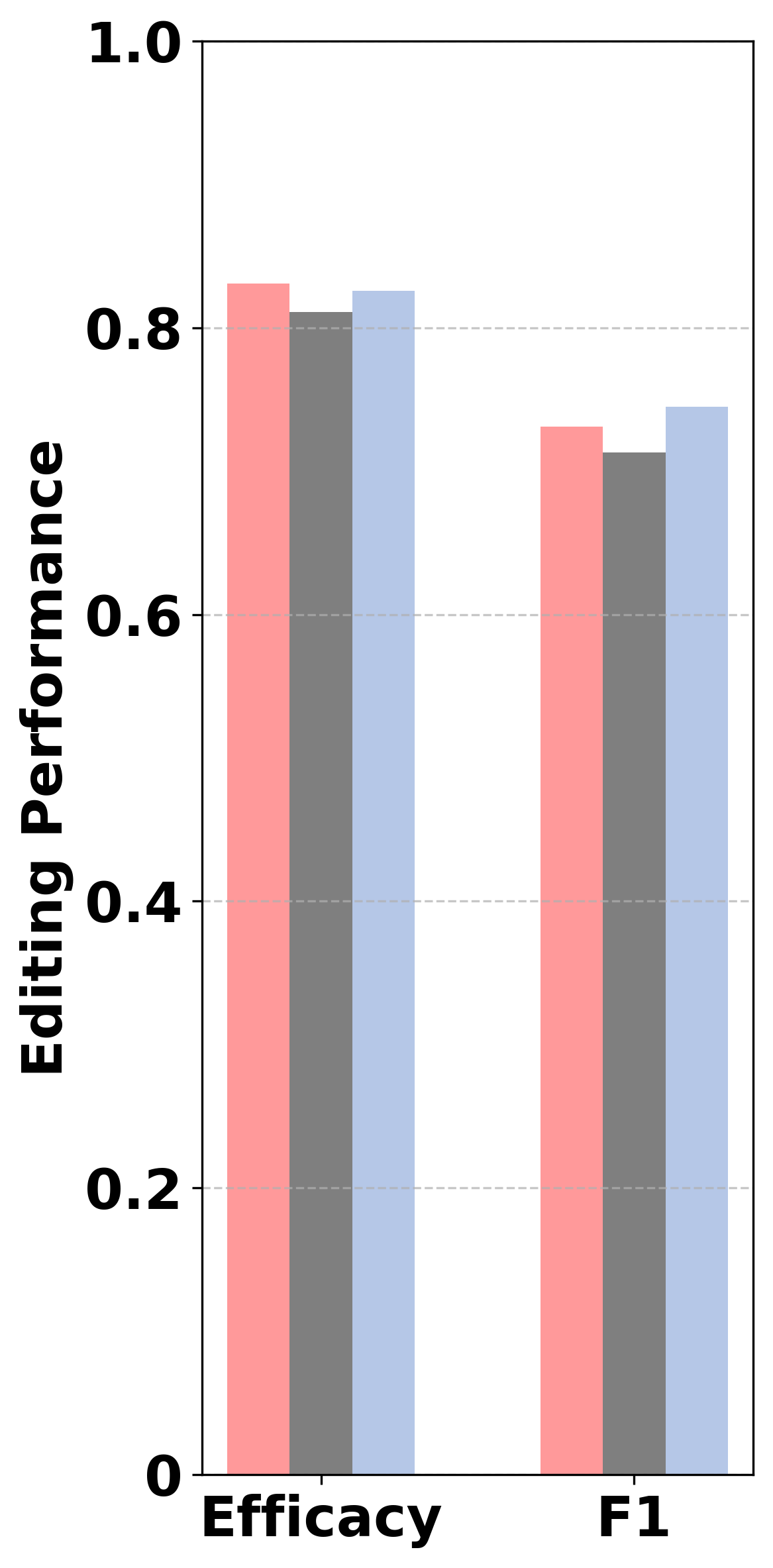}
\label{fig:b4_en}
}
\hfil
\subfloat[\scriptsize{ZH (bzsre)}]{\includegraphics[width=0.1\textwidth]{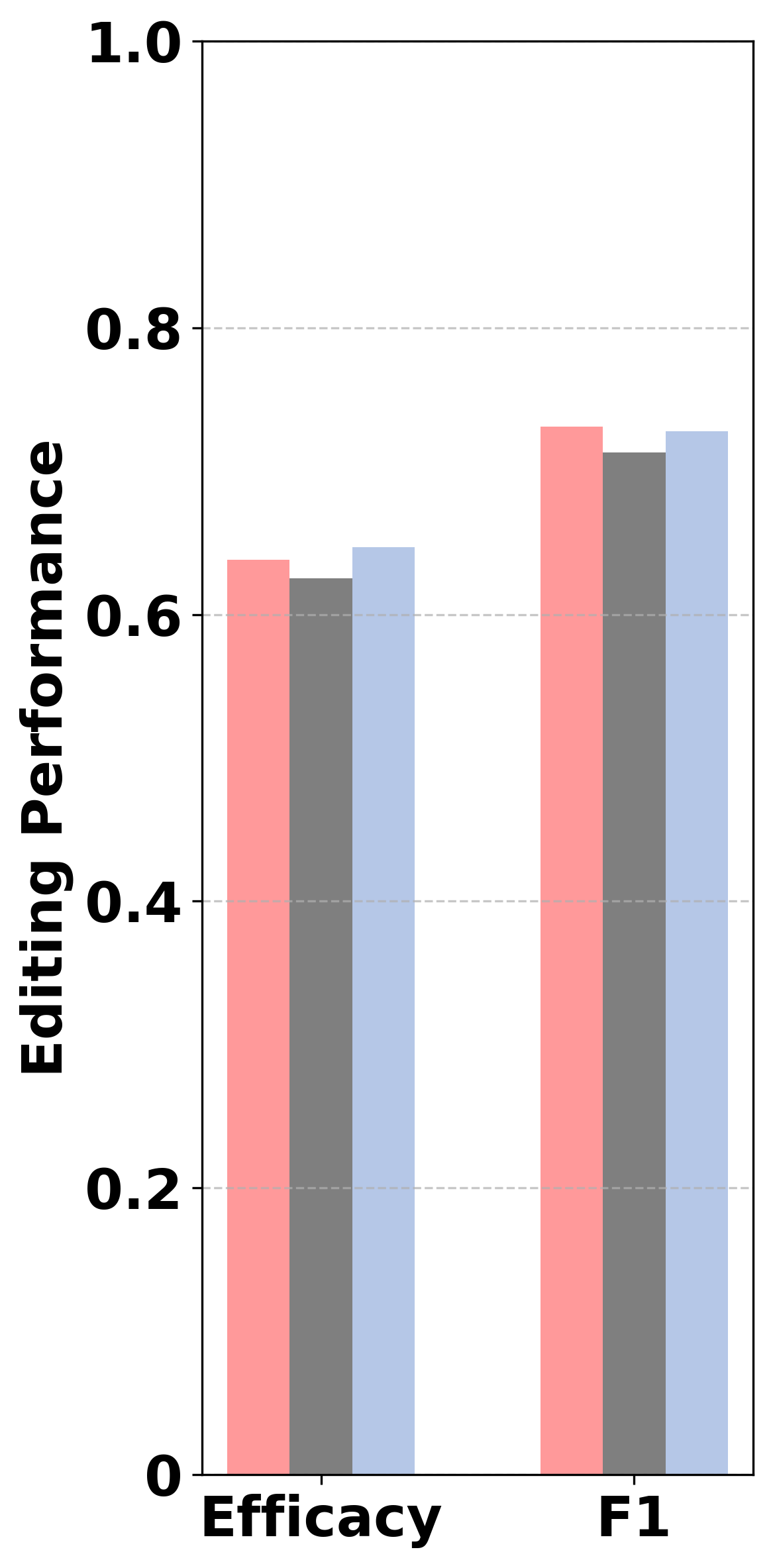}
\label{fig:b4_zhb}
}
\caption{Performance comparison for all languages between multilingual knowledge editing model and the state-of-the-art monolingual model. \redrect{}: LangEdit, \grayrect{}: AlphaEdit (Multi), \bluerect{}: AlphaEdit (Mono). }
\label{fig:mono_alpha_langedit}
\end{figure*}

\subsection{Per-Language Evaluation of Editing and Generalization}

To evaluate editing efficiency and multilingual generalization capabilities, we conducted comprehensive experiments using Llama3-8B as the backbone architecture across six languages. 

Figure~\ref{fig:radar_llama} presents radar charts illustrating the language-specific performance characteristics of LangEdit, AlphaEdit, and RECT.
Our analysis reveals two key findings:
(1) Our model consistently outperforms baselines across all evaluated languages, achieving state-of-the-art results in multilingual sequential knowledge editing. Specifically, it demonstrates improvements in editing accuracy (average +1.13) and F1 Scores on downstream tasks (average +3.32).
(2) We observe a notable disparity in performance gains between English and Spanish knowledge updates. While English updates show modest improvements (editing: +0.61, F1 Scores on downstream: average +3.39), Spanish updates yield significantly higher gains (editing: average +1.29, F1 Scores on downstream: average +9.00).
We attribute this discrepancy to the pretraining data imbalance in Llama3-8B, where the model's 
strong English generalization capacity leaves limited room for improvement through knowledge editing.
This finding aligns with recent studies on multilingual capacity scaling~\cite{conneau-etal-2020-unsupervised,fernandes2023scaling}, suggesting that editing effectiveness is inversely correlated with the pretraining data volume of the target language.

We further conduct in-depth analysis of knowledge sharing to explore whether a fact update in one language is accessible to other languages.
The results is illustrated in the Appendix~\ref{sec:knowledge_share}.

\subsection{Negative Interference in Multilingual Knowledge Editing}
\label{sec:mul}

To analyze negative interference, we evaluate Llama3-8B edited with knowledge in six languages using two editing datasets.
Editing efficacy is measured through the Efficacy Score, while the multilingual generalization ability is assessed by the averaged F1 Score. 
We formalize the magnitude of negative interference as the performance gap between the original AlphaEdit model - AlphaEdit (mono), designed for monolingual knowledge updates and a model designed for multilingual knowledge editing (AlphaEdit (multi) and LangEdit). 

Our experiments reveal significant negative interference in AlphaEdit (multi). 
As shown in Figure~\ref{fig:mono_alpha_langedit}, AlphaEdit (multi) exhibits substantial performance degradation in both Efficacy Score ($\Delta$ = +0.10 $\sim$ +3.27) and F1 Score ($\Delta$ = +3.20 $\sim$ +15.19) across all language pairs compared to the AlphaEdit (mono).
In contrast, our model demonstrates remarkable robustness to negative interference:
(1) It reduces the F1 Score disparity to ($\Delta$ = +0.20 $\sim$ +9.29) across languages, significantly lower than the ($\Delta$ = +3.20 $\sim$ +15.19) obtained by AlphaEdit.
(2) It surpasses AlphaEdit (mono) in Efficacy Score for English (+1.10), German (+3.01), Dutch (+2.58), and French (+3.95).

Without the proposed parameter updates the performance of multilingual sequential knowledge editing is lower than that of monolingual editing, confirming the effectiveness of the proposed null-space projection. Even for related languages (e.g., English-Dutch-German, French-Spanish), not integrating the null-space still degrades performance, reinforcing the effectiveness of our approach. 

\section{Related Work}

Knowledge editing approaches for large language models (LLMs) can be broadly categorized into two paradigms: parameter-modifying and parameter-preserving, depending on whether the model weights are updated. 
We focus on the parameter-modifying paradigm. 
Prior research~\cite{geva-etal-2021-transformer} has demonstrated that the MLP layers in Transformer models function as knowledge repositories, with specific neurons encoding editable factual associations.

Building on this insight, \citet{meng2022locating} introduced ROME (Rank-One Model Editing), a pioneering two-stage framework that first identifies the MLP layer storing the target knowledge and then injects a single knowledge edit through rank-one weight perturbations. This approach was later extended by MEMIT (Mass-Editing Memory in Transformer) \cite{mengmass}, which enhances scalability by enabling updates across multiple MLP layers to inject numerous knowledge edits. 

Subsequent works identified a critical challenge in sequential editing: \citep{gu-etal-2024-model} attributed degradation of general abilities to parameter perturbations and introduced RECT, a regularization technique constraining weight updates.
Concurrently, \citet{ma2024perturbation} established a mathematical connection between model degradation and the condition number~\cite{smith1967condition} of the edited weights, introducing PRUNE to impose condition number-based constraints on weight updates. 
The monolingual AlphaEdit~\cite{fang2024alphaedit} projected the parameter perturbation of the MLP weights into the null space of knowledge preserved in the LLMs.

The above methods have shown success in monolingual knowledge editing but are limited to one language.
To the best of our knowledge, we are the first work studying multilingual sequential knowledge editing, where the knowledge in different languages may interfere with each other.

Prior works~\cite{wang-etal-2024-cross,wang-etal-2024-retrieval,zhang-etal-2025-multilingual} focus on single knowledge editing~\cite{gu-etal-2024-model} instead of sequential knowledge editing. Single knowledge editing tests how well a model adapts to a single modification, while sequential knowledge editing checks whether the model can retain all previous edits and maintain overall performance after multiple consecutive changes.

\section{Conclusion}

We introduced the multilingual sequential knowledge editing task and identified the negative interference arising from parameter changes during sequential updates across languages. 
To address this, we proposed LangEdit, a novel framework employing null-space constrained optimization to isolate language-specific parameter updates while preserving the model's multilingual generalization capabilities. 
LangEdit achieves this by constructing "language safeguards", which prevent edits in one language-specific knowledge from adversely affecting performance in another, without the need for additional language-specific modules. 
LangEdit offers a technically sound method based on null-space projection, specifically adapted for the multilingual sequential setting with dynamic projections. 
Extensive experiments conducted across six languages and three large language models architectures demonstrate the effectiveness of LangEdit, establishing state-of-the-art performance in multilingual sequential knowledge editing.

\section*{Limitations}
While our study provides insight into multilingual sequential knowledge editing, three key limitations warrant further investigation:
(1) Our experiments are conducted on 6 to 8B parameter LLMs. Extending this analysis to larger architectures (e.g., 70B-scale models) could reveal scaling effects in multilingual knowledge editing.
Large models usually have more parameters, more complex structures, and have stronger multilingual capabilities. 
However, knowledge editing may face challenges, such as more parameters making it more difficult to locate specific knowledge areas, or more dispersed knowledge representation within the model.
(2) The current evaluation focuses on constrained editing scenarios. Future work should explore a broader range of downstream applications to assess real-world deployment viability.
(3) Our multilingual experiments are limited to 6 languages. Developing comprehensive multilingual benchmarks covering hundreds of languages would better test the boundaries of LangEdit and baselines.

\section*{Acknowledgements}
This research was funded by the CHIST-ERA projects ANTIDOTE (ERA-NET CHIST-ERA IV FET PROACT JTC 2019) and AIDAVA (EU HORIZON-HLTH-2021-TOOL-06-03). 

The computations described in this research were performed using the The Flemish Supercomputer Center Tier-1 HPC service~(https://www.vscentrum.be/). 

\bibliography{main}

\appendix

\section{Appendix}
\label{sec:appendix}
\subsection{The analysis of factual knowledge storage}
\label{sec:analysis_fact}

Assume that the factual knowledge stored in LLMs is represented in the $(s, r, o)$ format where $s$ denotes subjects, $o$ represents objects and $r$ is the relation between subjects and objects. 
\cite{meng2022locating} show that the MLP modules in the mid-layer encode subjects of the text in English and then generate outputs that retrieve updated knowledge from these layers.
However, it remains uncertain whether knowledge in other languages follows the same pattern as described in~\cite{meng2022locating}, necessitating further investigation.

\begin{figure*}[htbp]
\centering
\subfloat[\scriptsize{AIE of hidden states in EN}]{\includegraphics[width=0.32\textwidth]{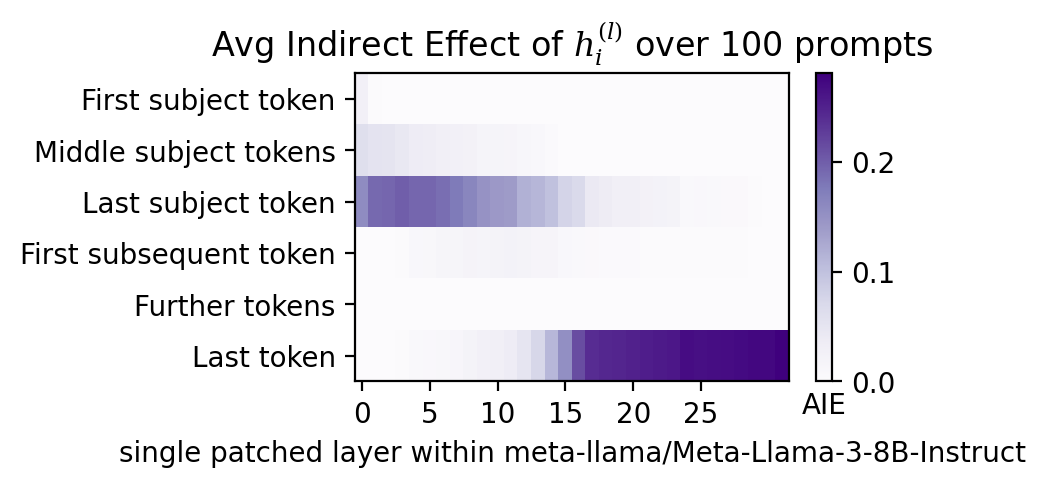}
}
\hfil
\subfloat[\scriptsize{AIE of MLPs in EN}]{\includegraphics[width=0.32\textwidth]{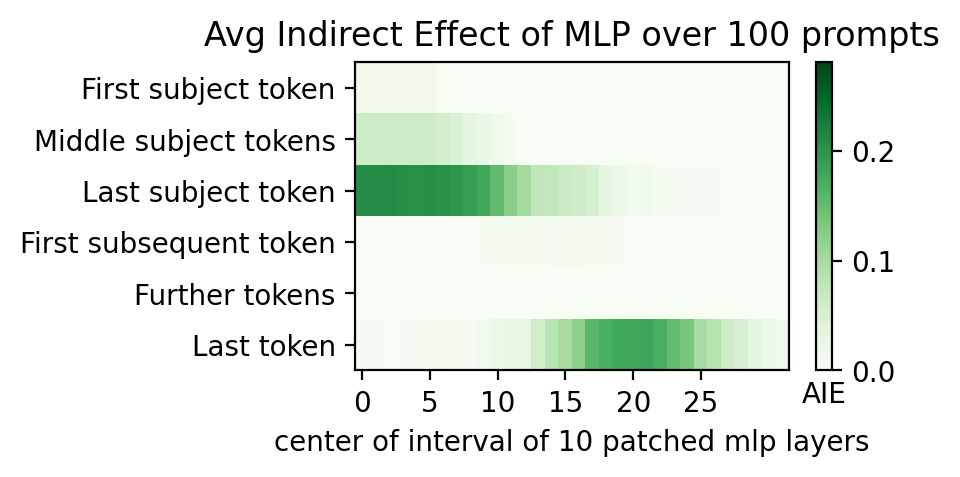}
}
\hfil
\subfloat[\scriptsize{AIE of Attns in EN}]{\includegraphics[width=0.32\textwidth]{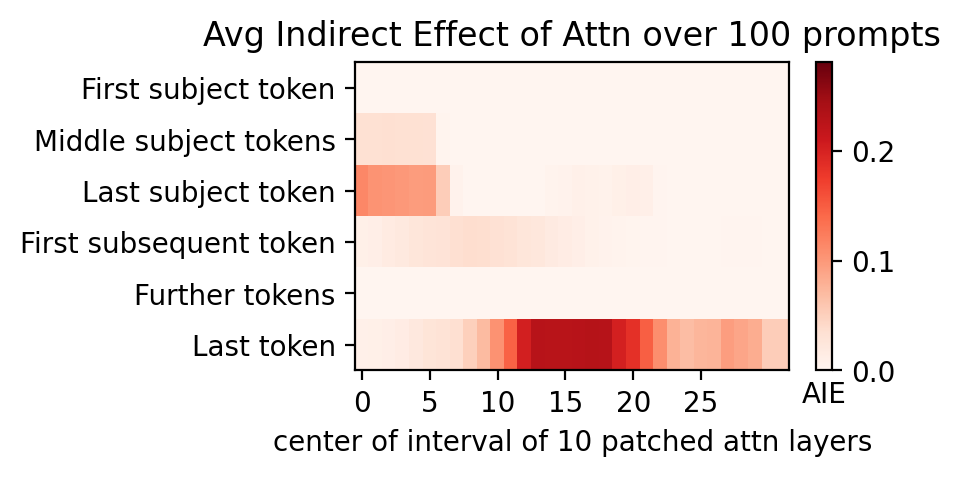}
}
\hfil
\subfloat[\scriptsize{AIE of hidden states in DE}]{\includegraphics[width=0.32\textwidth]{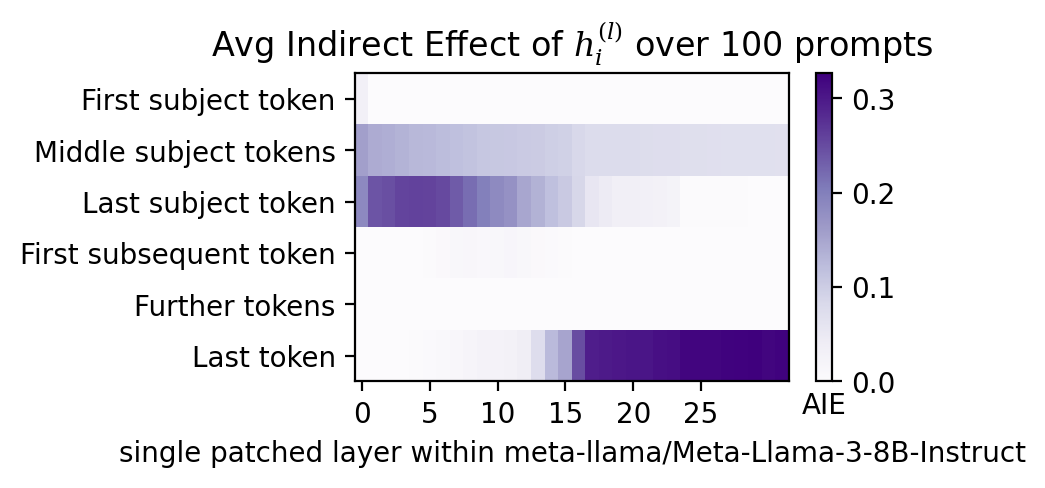}
}
\hfil
\subfloat[\scriptsize{AIE of MLPs in De}]{\includegraphics[width=0.32\textwidth]{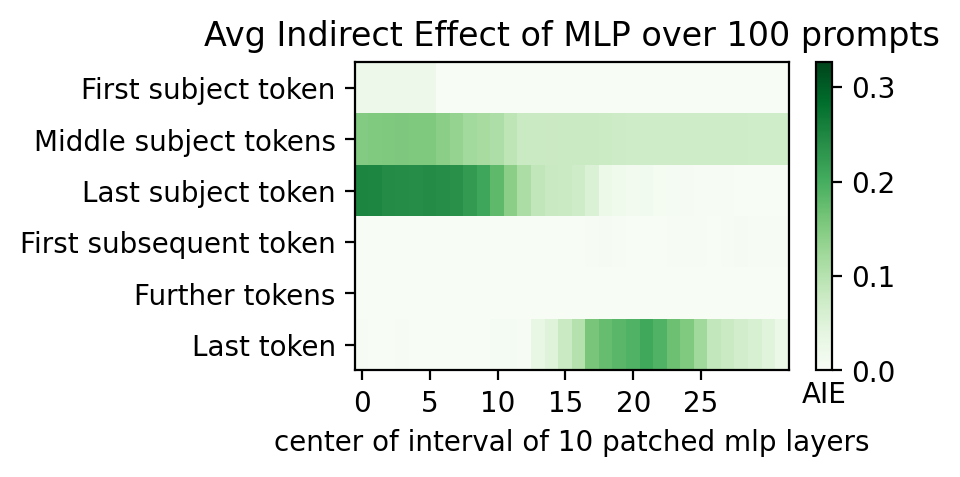}
}
\hfil
\subfloat[\scriptsize{AIE of Attns in DE}]{\includegraphics[width=0.32\textwidth]{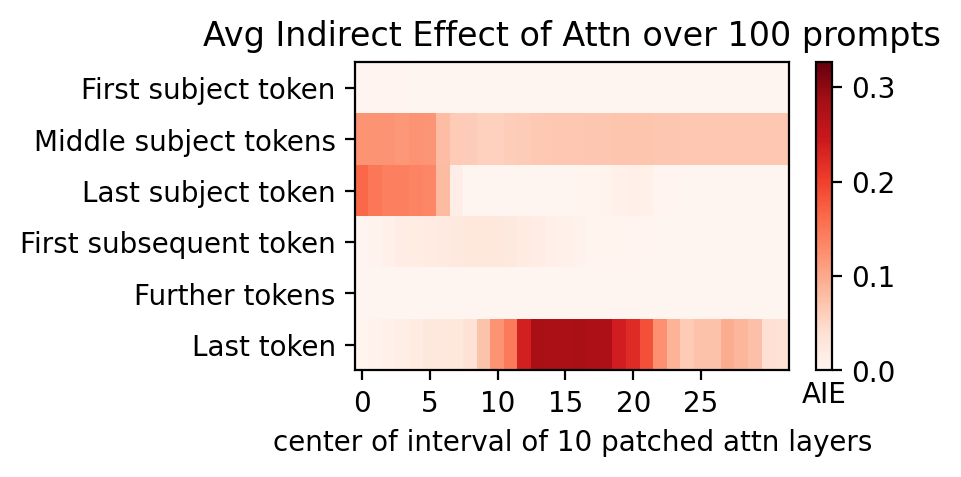}
}
\hfil
\subfloat[\scriptsize{AIE of hidden states in NL}]{\includegraphics[width=0.32\textwidth]{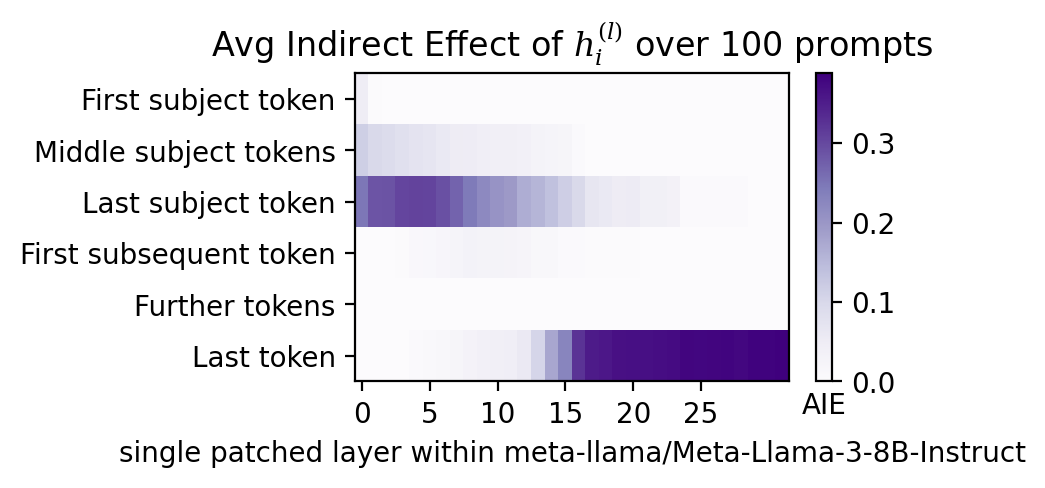}
}
\hfil
\subfloat[\scriptsize{AIE of MLPs in NL}]{\includegraphics[width=0.32\textwidth]{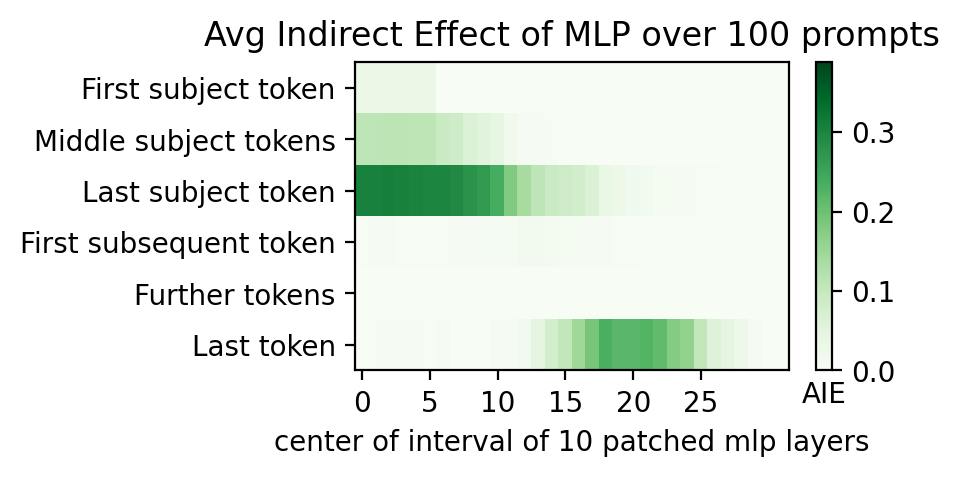}
}
\hfil
\subfloat[\scriptsize{AIE of Attns in NL}]{\includegraphics[width=0.32\textwidth]{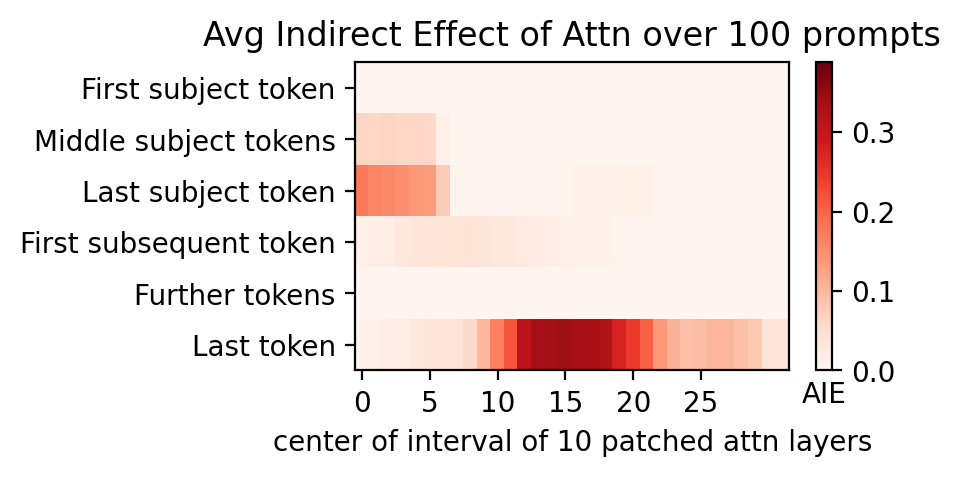}
}
\hfil
\subfloat[\scriptsize{AIE of hidden states in ES}]{\includegraphics[width=0.32\textwidth]{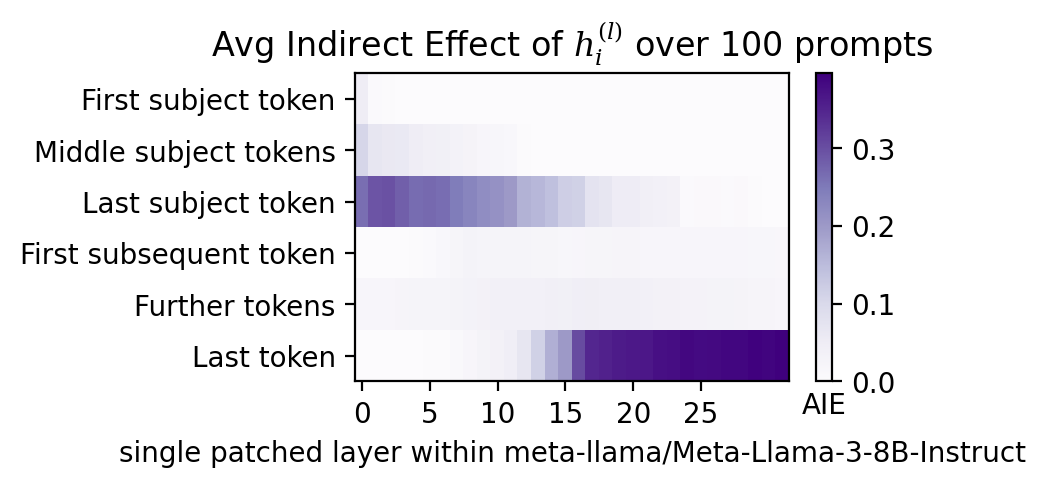}
}
\hfil
\subfloat[\scriptsize{AIE of MLPs in ES}]{\includegraphics[width=0.32\textwidth]{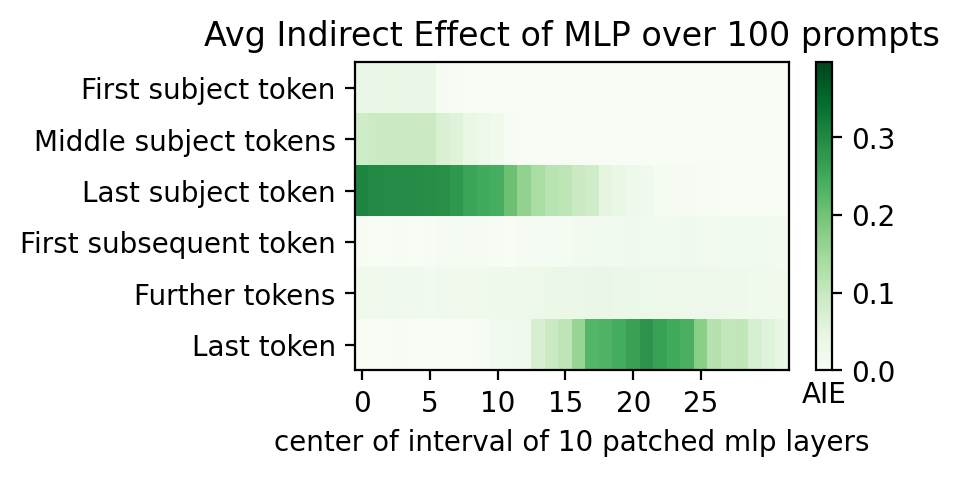}
}
\hfil
\subfloat[\scriptsize{AIE of Attns in ES}]{\includegraphics[width=0.32\textwidth]{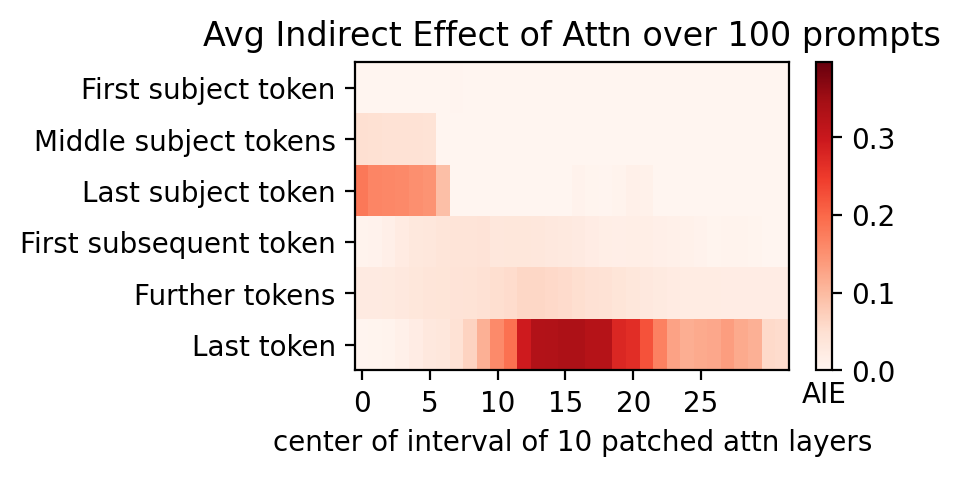}
}
\hfil
\subfloat[\scriptsize{AIE of hidden states in FR}]{\includegraphics[width=0.32\textwidth]{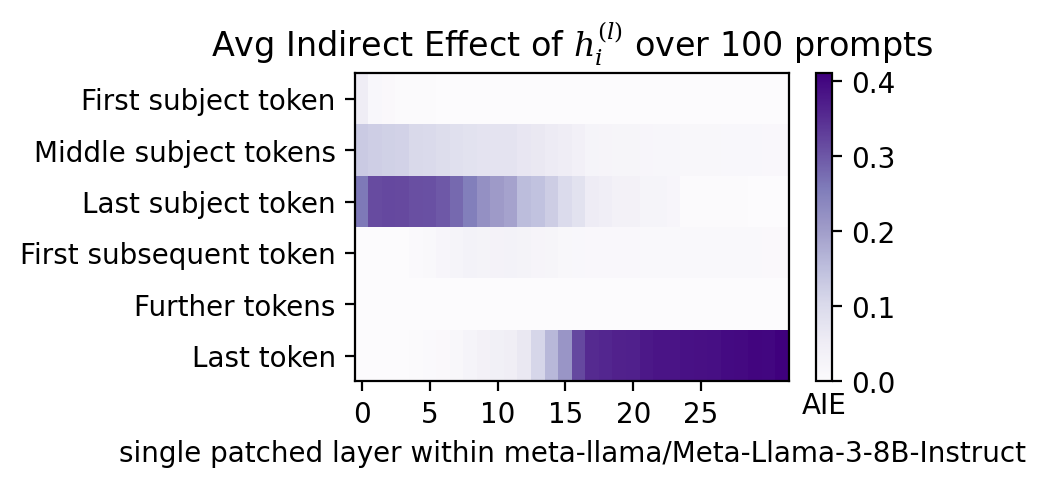}
}
\hfil
\subfloat[\scriptsize{AIE of MLPs in FR}]{\includegraphics[width=0.32\textwidth]{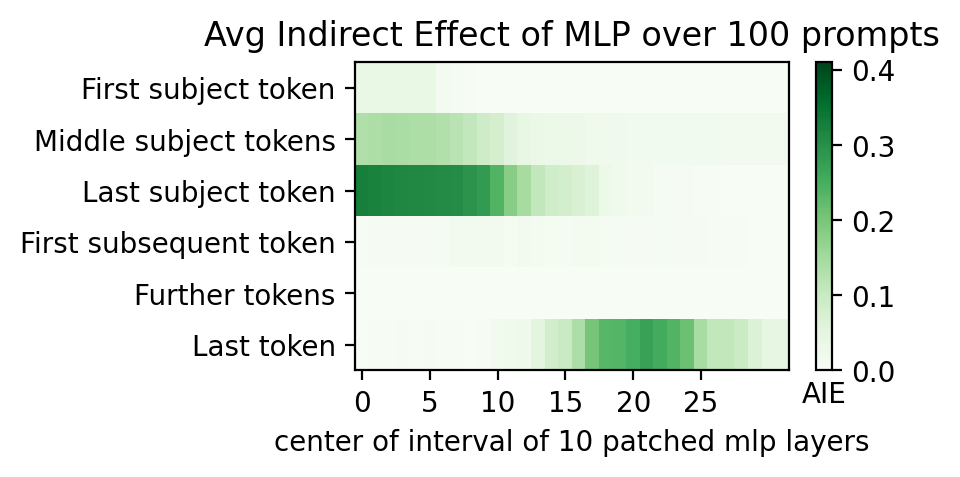}
}
\hfil
\subfloat[\scriptsize{AIE of Attns in FR}]{\includegraphics[width=0.32\textwidth]{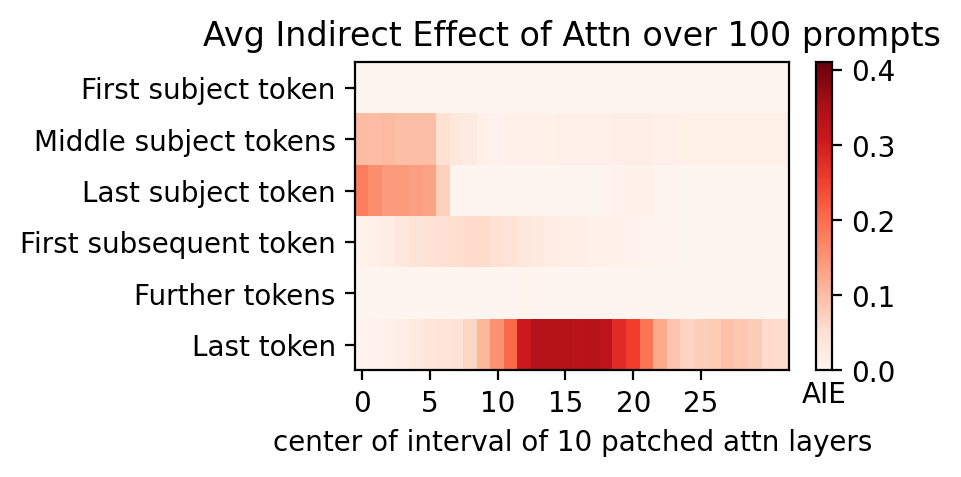}
}
\hfil
\subfloat[\scriptsize{AIE of hidden states in ZH}]{\includegraphics[width=0.32\textwidth]{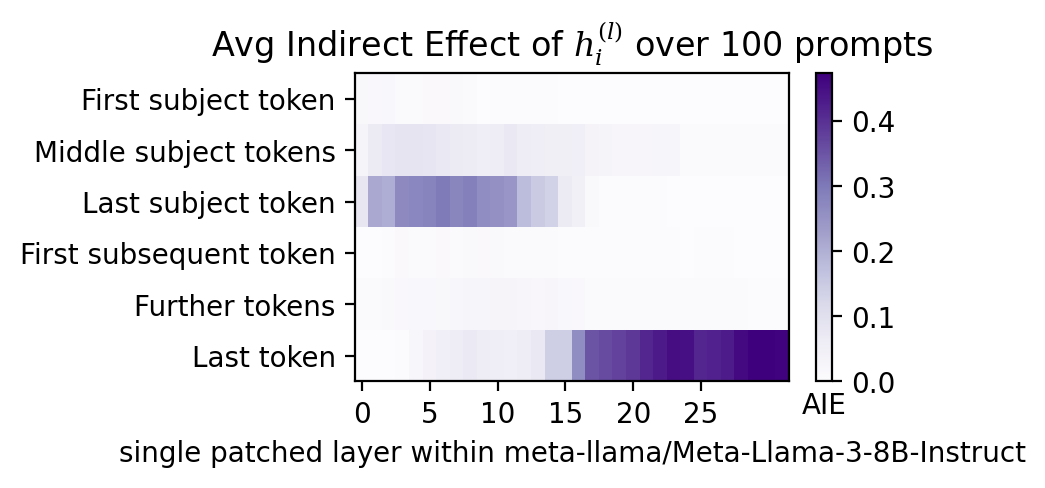}
}
\hfil
\subfloat[\scriptsize{AIE of MLPs in ZH}]{\includegraphics[width=0.32\textwidth]{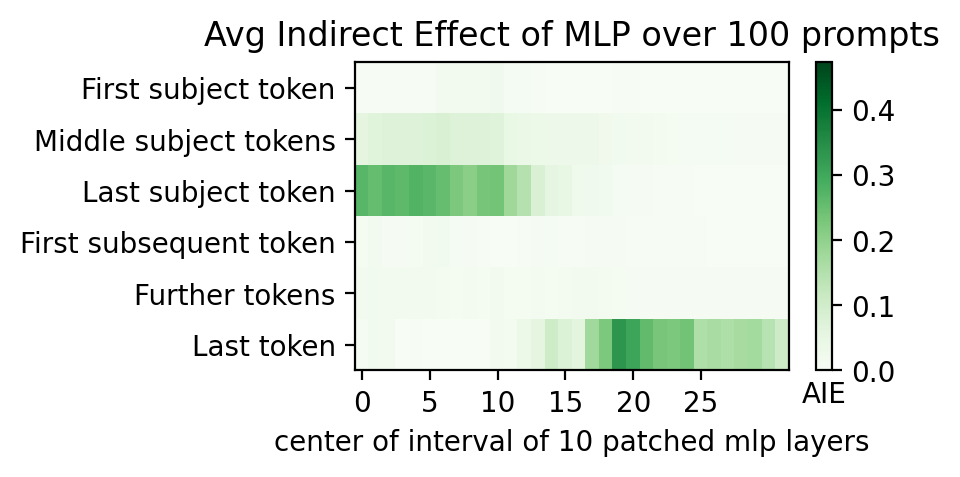}
}
\hfil
\subfloat[\scriptsize{AIE of Attns in ZH}]{\includegraphics[width=0.32\textwidth]{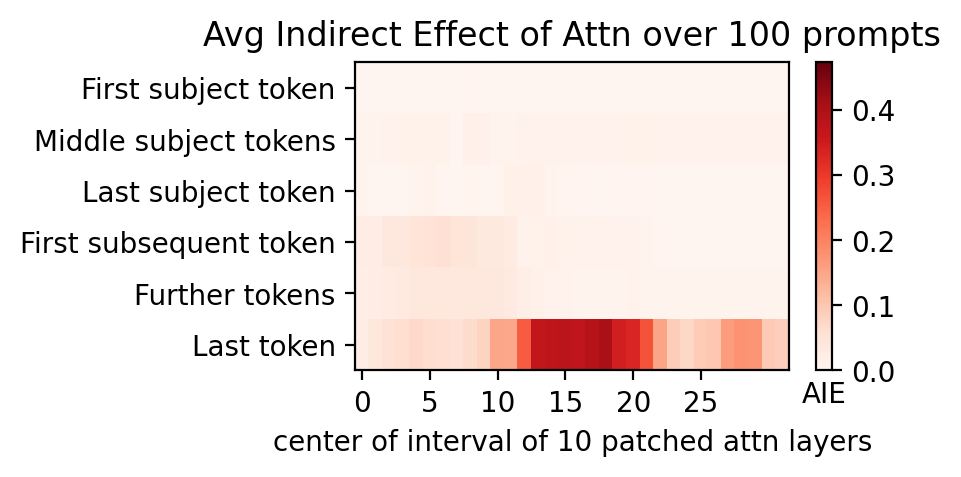}
}
\hfil
\caption{Average Indirect Effects of all model components~(Llama3-8B) of 100 factual knowledge in six languages.}
\label{fig:AIE_sum}
\end{figure*}

Firstly, we apply the causal tracing technique used in \cite{meng2022locating} to determine which hidden states have a causal impact on factual predictions by running the model multiple times, introducing interventions, and restoring specific states. 
The specific steps are as follows: 
\textbf{Clean Run}: The model processes the input without any interference, and the activation values of all hidden states are recorded. 
\textbf{Corrupted Run}: Noise (random perturbations) is introduced into the embeddings of the subject tokens to potentially cause the model to output incorrect factual predictions.
\textbf{Corrupted-with-Restoration Run}: Based on the corrupted run, specific hidden states are restored to observe whether these states can restore the model's correct prediction.
To quantify the causal contribution of each hidden state to factual predictions, we use average indirect effect~(AIE) to measure the importance of specific states.
Indirect Effect is the difference in predictions between the corrupted run and the corrupted-with-restoration run.
The first column of Figure~\ref{fig:AIE_sum} shows that strong causal states appear in early layers at the last token of the subject for all six languages. 
The second and third columns of Figure~\ref{fig:AIE_sum} suggest that the MLP contributes stronger causality in early layers compared to the attention module. 
The opinion obtained by~\cite{meng2022locating} is consistent with our finding, which is that the MLP modules in the mid-layers encode the subjects of knowledge in six languages and then generate the output of recalling memory objects.

\subsection{Experimental results of variant models for multilingual editing of the LLM}
\label{sec:var_lang}
We develop three variants and analyze their performance in Table~\ref{tab:full_model_editing——variant}:
\textbf{AlphaEdit (Translation): }
This variant edits only the English knowledge in the model using AlphaEdit and then translates the edited knowledge to other languages using Google Translate.
While it underperforms AlphaEdit in direct editing evaluation (likely due to translation errors), it achieves stronger performance on multilingual generalization tasks.
We assume that this is because focusing solely on English knowledge injection avoids interference from multilingual updates.
\textbf{AlphaEdit (Multilingual): }
We train the LLMs on multiple languages by first calculating knowledge updates for each language individually. Then, we combine all these updates by adding them together, forming unified updates that construct multilingual knowledge representations. 
Finally, we conduct knowledge editing by applying this representations to the weights of selected MLP layers in LLMs.
This approach underperforms compared to LangEdit across both editing accuracy and multilingual generalization, reinforcing the importance of multilingual sequential knowledge editing.
\textbf{LangEdit with shuffled language order of the knowledge): }
By randomizing the 
editing sequence for three times while preserving the content of the knowledge, we observe comparable performance to using the knowledge sequence without this shuffling.  This suggests that while sequential editing is crucial, the specific ordering of languages carries minimal importance.

\subsection{Package Version}
\label{sec:package}
Pytorch version is 2.3.0 and transformer is 3.9.1.

\subsection{Computational Analysis}
\label{sec:computation}
To evaluate the computational cost, we leverage the time per batch (100 edits) and memory cost. We conduct multilingual sequential knowledge editing on an NVIDIA A100-SXM4-80GB GPU. 
When we edit  Llama3-8B on the MzsRE dataset, Time per Batch (s) and Memory (GB) for all models are shown in Table~\ref{tab:llama3}.
When editing Qwen-7B on the MzsRE dataset, Time per Batch (s) and Memory (GB) for all models are shown in Table~\ref{tab:qwen}.
When editing GPT-J-6B on the MzsRE dataset, Time per Batch (s) and Memory (GB) for all models are shown in Table~\ref{tab:gptj}.

\begin{table}[ht]
\centering
\begin{tabular}{lrr}
\toprule
Method            & Time per Batch (s) & Memory (GB) \\
\midrule
FT                & 69                 & 36.8        \\
ROME              & 1804               & 36.7        \\
MEMIT             & 974                & 35.8        \\
PRUNE             & 1073               & 35.8        \\
RECT              & 991                & 35.8        \\
AlphaEdit & 1277              & 37.9        \\
LangEdit          & 1549               & 41.5        \\
\bottomrule
\end{tabular}
\caption{Computational cost of knowledge methods when editing the Llama3-8B.}
\label{tab:llama3}
\end{table}

\begin{table}[ht]
\centering
\begin{tabular}{lrr}
\toprule
Method            & Time per Batch (s) & Memory (GB) \\
\midrule
FT                & 68                 & 28.1        \\
ROME              & 1509               & 30.9        \\
MEMIT             & 792                & 35.6        \\
PRUNE             & 809                & 35.6        \\
RECT              & 845                & 35.6        \\
AlphaEdit & 963               & 34.7        \\
LangEdit          & 1098               & 37.9        \\
\bottomrule
\end{tabular}
\caption{Computational cost of knowledge methods when editing the GPT-J-6B.}
\label{tab:gptj}
\end{table}

\begin{table}[t!]
\centering
\begin{tabular}{lrr}
\toprule
Method            & Time per Batch (s) & Memory (GB) \\
\midrule
FT                & 69                 & 32.6        \\
ROME              & 1183               & 36.5        \\
MEMIT             & 450                & 35.7        \\
PRUNE             & 479                & 35.7        \\
RECT              & 485                & 35.7        \\
AlphaEdit & 524               & 34.4        \\
LangEdit          & 793                & 38.3        \\
\bottomrule
\end{tabular}
\caption{Computational cost of knowledge methods when editing the Qwen-7B.}
\label{tab:qwen}
\end{table}

\begin{table*}[t!]
    \centering
    {\small
    \renewcommand{\arraystretch}{1.2}
    \setlength{\tabcolsep}{1pt}
    \begin{tabular}{l |ccc|c|ccc|c} 
        \toprule
        \multicolumn{1}{c}{\textbf{Methods}} & \multicolumn{3}{c}{\textbf{mzsre}} & XTREME $\ddag$ & \multicolumn{3}{c}{\textbf{bzsre}} & XTREME $\dag$\\
        \cmidrule(lr){2-4} \cmidrule(lr){6-8}
         \multicolumn{1}{c}{} & \multicolumn{1}{c}{\textbf{Efficacy$\uparrow$}} & \textbf{Generality$\uparrow$} & \textbf{Specificity$\uparrow$} & {\textbf{F1$\uparrow$}} & {\textbf{Efficacy$\uparrow$}} & \textbf{Generality$\uparrow$} & \textbf{Specificity$\uparrow$} & \textbf{F1$\uparrow$} \\
        \midrule
         AlphaEdit~(translation) & 63.01 & 55.57 & 24.24 & 75.53 & 53.91 & 47.74 & 24.66 & 74.51 \\
         AlphaEdit~(multilingual) & 45.35 & 37.53 & 11.23 & 14.53 & 39.15 & 22.21 & 9.98 & 4.66 \\
         LangEdit~(shuffle order) & 81.69 & 76.75 & 32.77 & 65.37 & 72.97 & 67.11 & 30.09 & 73.01 \\
         LangEdit & 82.54 & 77.53 & 31.90 & 66.24 & 73.18 & 66.95 & 31.11 & 73.14 \\
        \bottomrule
    \end{tabular}
    }
    \caption{Performance of variant models for multilingual editing using Llama3-8B and evaluated on the  mzsre and bzsre datasets. XTREME$\ddag$ represents  average F1 Scores on XTREME tasks after training on the mzsre dataset; XTREME$\dag$ denotes the average F1 Scores after training on the bzsre dataset.  All baselines are adapted for multilingual sequential knowledge editing.}
    \label{tab:full_model_editing——variant}
\end{table*}

\subsection{The Analysis on Knowledge Sharing}
\label{sec:knowledge_share}
To explore whether a fact expressed in one language remains consistent across other languages, we conduct the following experiment, illustrated with the example below.
Imagine injecting new facts into an English-Spanish bilingual model: 
\begin{itemize}
    \item We edit the LLM with English data (e.g., "Carl Sagan → worked at BBC"). 
\item We keep the original Spanish data unchanged. 
\item If the model correctly answers the Spanish query 'Dónde trabajó Carl Sagan?' ('Where did Carl Sagan work?')—despite never having seen this edit in Spanish—it demonstrates successful cross-lingual transfer enabled by LangEdit's architecture.

\end{itemize}

The experimental results for exploring cross-lingual knowledge transfer with queries in the test language that have never been seen in the edits are shown in Table~\ref{tab:results_fr_es} and Table~\ref{tab:results_en_de_nl}.
From Table~\ref{tab:results_fr_es}, we can observe that the evaluation scores of LangEdit (Spanish) and LangEdit (French) are higher than the Llama3 (Spanish) and Llama3 (French). 
The same phenomenon appears in Table~\ref{tab:results_en_de_nl}. 
This indicates that the edited knowledge in a certain language can propagate to other languages when using LangEdit.

\begin{table*}[t!]
\centering
\begin{tabular}{lcccc}
\toprule
Models & Test language & Efficacy & Generality & Specificity \\
\midrule
LangEdit & French & 56.41 & 53.91 & 30.54 \\
Llama3 (unedited) & French & 31.51 & 30.90 & 30.21 \\
\hline
LangEdit & Spanish & 55.89 & 53.84 & 31.54 \\
Llama3 (unedited) & Spanish & 31.71 & 31.62 & 31.33 \\
\bottomrule
\end{tabular}
\caption{Experimental results of cross-lingual knowledge transfer when editing Llama3-8B in French and Spanish.}
\label{tab:results_fr_es}
\end{table*}

\begin{table*}[t!]
\centering
\begin{tabular}{lcccc}
\toprule
Models & Test language & Efficacy & Generality & Specificity \\
\midrule
LangEdit & English & 59.02 & 57.37 & 31.14 \\
Llama3 (unedited) & English & 31.03 & 30.92 & 30.97 \\
\hline
LangEdit & German & 58.12 & 56.30 & 30.16 \\
Llama3 (unedited) & German & 30.15 & 29.71 & 29.95 \\
\hline
LangEdit & Dutch & 56.33 & 53.31 & 30.82 \\
Llama3 (unedited) & Dutch & 30.62 & 30.14 & 30.37 \\
\bottomrule
\end{tabular}
\caption{Experimental results of cross-lingual knowledge transfer when editing Llama3-8B in English, German and Dutch.}
\label{tab:results_en_de_nl}
\end{table*}

\subsection{Experimental Results on the MLaKE Dataset}
\label{sec:mlake}
MLaKE is a multilingual benchmark designed to evaluate the performance of knowledge editing methods in large language models across different languages and reasoning complexities. 
It contains 4072 multi-hop and 5360 single-hop question-answer pairs in five languages: English, Chinese, Japanese, French, and German. 
Each instance is based on fact chains aligned from Wikipedia, covering both shallow and complex reasoning paths. MLaKE enables systematic evaluation of cross-lingual transferability, multi-hop reasoning, and the limitations of current multilingual knowledge editing approaches.
The MLaKE dataset does not provide subject items for the knowledge triplets. 
We conduct multilingual sequential knowledge editing on the MLaKE dataset in three languages (English, German and French), where we have manually annotated the subject items for a small subset~(100 examples for each language). Experimental results on the MLaKE dataset  are shown in the Table~\ref{tab:model_comparison—_mlake}.

\begin{table*}[h]
\centering
\begin{tabular}{|l|cc|cc|}
\hline
\textbf{Models} & \multicolumn{2}{c|}{\textbf{Single-hop}} & \multicolumn{2}{c|}{\textbf{Multi-hop}} \\
                & \textbf{Efficacy} & \textbf{XTREME (F1)} & \textbf{Efficacy} & \textbf{XTREME (F1)} \\
\hline
AlphaEdit (adapted)  & 90.62 & 69.96 & 88.48 & 68.17 \\
LangEdit             & 91.44 & 70.89 & 90.05 & 69.09 \\
\hline
\end{tabular}
\caption{Comparison of models on Single-hop and Multi-hop tasks using Efficacy and XTREME (F1) metrics.}
\label{tab:model_comparison—_mlake}
\end{table*}

\end{document}